\documentclass{article}

\usepackage{arxiv}

\usepackage[utf8]{inputenc}
\usepackage[T1]{fontenc}
\usepackage{hyperref}
\usepackage{url}
\usepackage{amsmath,amssymb,amsfonts}
\usepackage{booktabs}
\usepackage{multirow}
\usepackage{graphicx}
\usepackage{overpic}
\usepackage{xcolor}
\usepackage{algorithm}
\usepackage{algpseudocode}
\usepackage{microtype}
\usepackage[round]{natbib}

\usepackage{amsmath,amsfonts,bm}

\def\eqref#1{equation~\ref{#1}}

\def\1{\bm{1}}

\DeclareMathAlphabet{\mathsfit}{\encodingdefault}{\sfdefault}{m}{sl}
\SetMathAlphabet{\mathsfit}{bold}{\encodingdefault}{\sfdefault}{bx}{n}

\newcommand{\Mname}{FSC}

\title{Frequency-Step Cache: Separating Synchronization and Repair for Efficient Autoregressive Video Diffusion}

\renewcommand{\shorttitle}{Frequency-Step Cache}

\author{
  Kerry Xu \quad
  Param Misra \quad
  Zachary Stenglein \quad
  Athri Karunamurthy \quad
  Fanjiang Ye \quad
  Yuke Wang \\
  Rice University \\
  \texttt{yuke.wang@rice.edu}
}

\begin{document}

\maketitle

\vspace{-0.8em}
\begin{abstract}
Autoregressive (AR) video generation makes long-horizon visual synthesis possible,
but every generated chunk requires repeated forward passes through a large
diffusion transformer (DiT).
Training-free caching reduces this cost by reusing intermediate computation.
However, existing cache controllers often couple two decisions:
\emph{when} the cache is synchronized exactly and \emph{how much}
approximate computation is spent in between.
Our characterization suggests that these decisions should be separated.
For reactive staleness signals, exact synchronization should be scheduled
before measured drift accumulates, while measured drift should decide only
how much repair to spend between synchronizations.
Based on this separation, we introduce \textbf{Frequency-Step Cache (FSC)},
a training-free controller that performs exact synchronization at
full-computation anchors on a dense-to-sparse schedule and adaptive repair
between them.
A lightweight latent-space risk combining frequency-resolved drift with token
age routes each non-anchor step from whole-step replay to selective token repair
or full repair without changing the synchronization schedule.
We further show on SkyReels-V2 that whole-step reuse provides the main
acceleration, while selective token computation is most effective as fidelity
repair.
In a controlled ablation, using one reactive signal for both synchronization
and repair reduces fidelity at matched latency, with the separation advantage
also observed on FramePack and Matrix-Game.
Across multiple AR video backbones, FSC provides favorable speed--quality
trade-offs.
On SkyReels-V2, \Mname{}-Speed reaches a $2.36\times$ speedup, above
MotionCache's $2.275\times$, while improving SSIM by $0.019$ and PSNR by
$0.5$~dB.
At MotionCache's speed, \Mname{}-Quality improves SSIM by $0.031$ and
PSNR by $1.0$~dB, while lowering LPIPS by $0.024$.
Without retuning, \Mname{}-Quality transfers to the 14B SkyReels-V2 model
at $2.12\times$ speedup, compared with $1.79\times$ for MotionCache.
\end{abstract}
\begin{center}
    \includegraphics[width=0.98\textwidth]{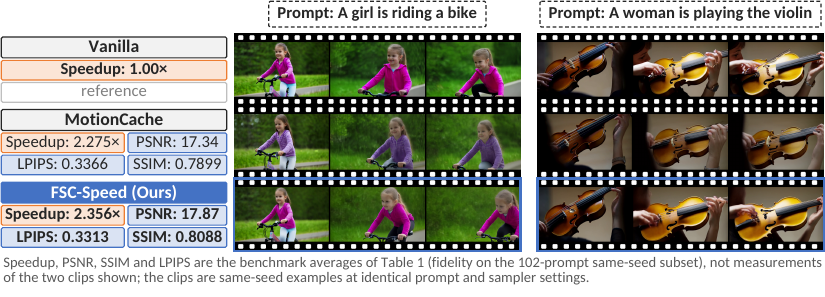}
\end{center}

\vspace{0.3em}
{\small
\textbf{Visual quality comparison: \Mname{}-Speed exhibits smaller deviations from the vanilla trajectory than MotionCache~\citep{xu2026motioncache}.}
In particular, \Mname{} better preserves the cloth color in the left example and the violin's appearance in the right example.
}
\vspace{1em}

\section{Introduction}
\label{sec:intro}

Autoregressive (AR) video generation has emerged as a practical paradigm for
long-horizon visual synthesis, generating videos sequentially by conditioning
each new chunk on previously generated content
~\citep{chen2024diffusionforcing,yin2025causvid,chen2025skyreelsv2,teng2025magi1}.
This chunk-wise formulation makes long-video generation possible while keeping
attention context and memory use bounded.
However, each chunk still requires tens of denoising steps, which refine a
noisy video representation through repeated evaluations of a large diffusion
transformer~\citep{peebles2023dit}.
Training-free caching reduces this cost by reusing intermediate outputs across
nearby steps~\citep{ma2024deepcache,selvaraju2024fora,liu2024teacache}.
Yet redundancy varies across steps and video content, making it difficult to
decide when reuse remains accurate
\citep{zou2025toca,ma2026flowcache,xu2026motioncache,liu2025freqca}.
Reuse introduces approximation errors that can accumulate across steps and
chunks~\citep{cui2026dpcache,peng2026ertacache,nan2026arcache}.
Existing cache controllers often use reactive reuse signals to govern both
when exact computation occurs and how much approximation is allowed in between.
These two roles operate differently: synchronization establishes the reference state for subsequent reuse, whereas
repair responds to staleness observed after reuse.
Cache control therefore involves two distinct decisions.
\textbf{(1) When should the cache be fully refreshed?}
We call this \emph{exact synchronization}: full computation updates cached
outputs and establishes a reference for measuring subsequent changes in the
video representation.
\textbf{(2) How much should be recomputed between refreshes?}
We call this \emph{adaptive repair}: choosing how much computation to perform
based on observed changes that indicate inaccurate reuse.

Our characterization motivates separating these decisions
(Sec.~\ref{sec:characterization}).
Reuse accuracy varies strongly by step, favoring scheduled synchronization
(Fig.~\ref{fig:characterization}(a)).
Approximate synchronization leaves errors in the cached outputs reused by
later steps, while change-based measurements report deviations only after they
occur (Fig.~\ref{fig:characterization}(d)).
We therefore schedule exact synchronization independently of these
measurements and use them only to allocate repair.
In a controlled ablation, a purely reactive variant has lower fidelity at
matched latency on all eight development prompts (Sec.~\ref{sec:ablation_separation}).
Between synchronizations, repair must balance accuracy against execution cost.
On SkyReels-V2, selectively computing video tokens, the model's units of video
representation, yields much less speedup than the computation reduction
suggests (Fig.~\ref{fig:characterization}(c)).
This favors reusing an entire step's cached outputs for acceleration and
selective computation for repair.
To guide repair, a lightweight score based on changes in the video representation tracks reuse error.
The score reaches a Spearman rank correlation of $0.893$ with reuse error measured against full computation,
as shown in Fig.~\ref{fig:oracle_rho}.

Guided by these findings, \textbf{Frequency-Step Cache (\Mname{})} schedules
full-computation steps, called \emph{anchors}, progressively farther apart
(Sec.~\ref{sec:framework_schedule}).
Between anchors, it reuses cached outputs for a whole step or repairs them by
recomputing varying numbers of tokens or performing full computation
(Sec.~\ref{sec:framework_routing}).
The repair score combines low- and high-frequency changes, representing coarse structure
and fine detail respectively, with token age, the steps since each token's last update
(Sec.~\ref{sec:design_risk}).
These measurements control repair, never the anchor schedule.
Our contributions are:
(i) separating scheduled synchronization from measurement-driven repair,
outperforming coupled control at approximately matched latency on SkyReels-V2,
FramePack, and Matrix-Game;
(ii) showing on SkyReels-V2 that whole-step reuse supplies acceleration while
selective token computation is most effective as fidelity repair; and
(iii) \Mname{}, a training-free controller that combines scheduled exact
anchors with lightweight frequency- and age-aware adaptive repair.

Experiments across multiple AR video backbones demonstrate favorable
speed--quality trade-offs and consistently support the separation principle.
On SkyReels-V2, \Mname{}-Speed reaches a $2.36\times$ speedup, above MotionCache's $2.275\times$, while
improving same-seed SSIM by $0.019$ and PSNR by $0.5$~dB, with comparable VBench quality. At MotionCache-matched speed,
\Mname{}-Quality improves SSIM by $0.031$, PSNR by $1.0$~dB, and lowers LPIPS by $0.024$.

\section{Related Work}
\label{sec:background}

\paragraph{Autoregressive Video Generation.}
Recent AR video models generate long videos as sequences of causally
conditioned frames or chunks with diffusion backbones
~\citep{chen2024diffusionforcing,yin2025causvid}, often trained with flow
matching~\citep{chen2025skyreelsv2,teng2025magi1}.
This chunk-wise formulation enables long-horizon generation under bounded
context and memory, but each chunk still requires repeated denoising
computation.

\paragraph{Training-Free Diffusion Caching.}
Feature caching reduces repeated denoising computation by reusing
activations or residuals across nearby steps, using fixed schedules
(FORA~\citep{selvaraju2024fora}, $\Delta$-DiT~\citep{chen2024deltadit}),
adaptive timing (TeaCache~\citep{liu2024teacache},
DiCache~\citep{bu2025dicache}), or forecasting
(TaylorSeer~\citep{liu2025taylorseer}).
Token-wise methods instead combine reuse with selective recomputation:
ToCa~\citep{zou2025toca} uses a predetermined refresh cycle with attention- and age-based token ranking,
DuCa~\citep{zou2024duca} alternates reuse and token recomputation, and
ProCache~\citep{cao2026procache} uses an offline-searched reuse schedule.
ToCa studies token-wise caching in diffusion transformers but not chunk-autoregressive long-video cache control.
It also fixes both the refresh cycle and token-recomputation budget a priori; attention and cache age
rank which tokens receive that budget, whereas \Mname{} adapts the repair amount online while keeping
synchronization scheduled.

\paragraph{Caching for Autoregressive Video.}
FlowCache~\citep{ma2026flowcache} assigns
chunk-level reuse policies, MotionCache~\citep{xu2026motioncache} introduces
motion-aware token adaptivity, and ARCache~\citep{nan2026arcache} determines
whole-step reuse from history-token changes.
SCOPE~\citep{scope2026} expands the inter-step action space through prediction and
selective frame computation.
These methods adapt reuse to temporal redundancy, but do not explicitly formulate exact-synchronization timing and
intermediate repair amount as separate control decisions.

\paragraph{Frequency-Aware Cache Control.}
Frequency-aware methods use spectral information to schedule
whole-step refreshes (SeaCache~\citep{chung2026seacache}) or manipulate
frequency components (FreqCa~\citep{liu2025freqca},
SpectralCache~\citep{li2026spectralcache}).
PreciseCache~\citep{wang2026precisecache} uses low-frequency differences to decide when denoising computation can be skipped.
\Mname{} instead keeps exact synchronization timing independent of measured staleness: synchronization is scheduled,
while frequency-resolved drift and cache age determine only inter-anchor repair.

\section{Characterization and Observations}
\label{sec:characterization}

\newcommand{\tightparagraph}[1]{%
  \vspace{-0.45em}%
  \paragraph{#1}%
}

We first characterize where redundancy exists and where naive caching fails in controlled profiling experiments
(App.~\ref{sec:appendix_char_protocol}).
The resulting observations motivate the design; full-scale autoregressive ablations of the deployed controller
(Sec.~\ref{sec:experiments}) reproduce the relevant orderings.

\begin{figure*}[!t]
\centering
\includegraphics[width=\linewidth]{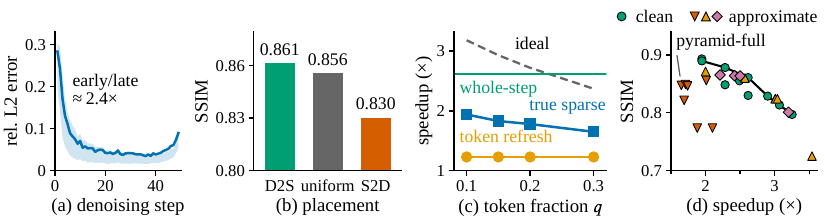}
\vspace{-10pt}
\caption{
\textbf{Three limitations of naive caching: reuse error depends on the step, token-selective repair provides limited
speedup, and approximate anchors rarely help (Observations 1--3).}
SkyReels-V2, controlled profiling setting (App.~\ref{sec:appendix_char_protocol}): (a) 14B; (b)--(d) 1.3B.
Anchor counts differ because each panel is a separate controlled study.
\textbf{(a)} Relative-$L_2$ error from reusing each block's previous-step residual
(band: interquartile range over blocks and prompts).
\textbf{(b)} SSIM vs.\ same-seed vanilla for three placements of $7$ ($8$ for uniform) exact anchors:
dense-to-sparse (D2S, $0.861$) outperforms uniform ($0.856$) and sparse-to-dense (S2D, $0.830$).
\textbf{(c)} Speedup vs.\ token fraction $q$.
True sparse kernels reach $1.94\times$ at $q{=}0.10$, far below their compute-proportional $2.72\times$; the
dense-forward token-refresh reference stays at $\approx1.23\times$ by design (dashed: its ideal).
Whole-step D2S reuse reaches $2.62\times$.
\textbf{(d)} SSIM vs.\ speedup for anchor policies (``clean'' circles: exact anchors; ``approximate'':
$\triangledown$ pyramid kernels, $\triangle$ block-skip gates, $\Diamond$ other variants).
Of $17$ approximate variants, only three block-skip gates near $3.0\times$ lie above the exact-anchor frontier
(black line), by at most $0.008$ SSIM.
The pyramid-full variant has substantially lower SSIM than the clean anchor baseline.
}
\label{fig:characterization}
\vspace{-0.5em}
\end{figure*}

\tightparagraph{Observation 1: Cacheability is strongly step-dependent.}
Reusing the previous step's residuals at early steps incurs approximately $2.4\times$ the error of reuse at late
steps (Fig.~\ref{fig:characterization}(a)).
Under a comparable anchor budget, dense-to-sparse placement yields the highest mean SSIM
(Fig.~\ref{fig:characterization}(b); Sec.~\ref{sec:ablation_schedule}).
\emph{Design consequence:} anchors should be dense early and increasingly sparse later
(Sec.~\ref{sec:framework_schedule}).

\tightparagraph{Observation 2: Sparse token computation gives limited speedup on SkyReels-V2.}
Fig.~\ref{fig:characterization}(c) profiles true sparse kernels, which recompute a subset of tokens per step: at
$q{=}0.10$ they reach $1.94\times$ against a compute-proportional $2.72\times$, mainly because full-context key/value
handling and launch overhead do not shrink with token fraction.
Their stale keys and values also cost $\approx0.024$ SSIM relative to a dense-forward token-refresh quality reference
(its $\approx1.23\times$ is fixed by the full forward; App.~\ref{sec:appendix_char_o2}).
Whole-step dense-to-sparse reuse reaches $2.62\times$.
\Mname{}'s deployed token repair is a gathered-batch implementation whose cost is measured end to end.
\emph{Design consequence:} on this backbone, whole-step reuse should supply the acceleration, and token computation
should primarily serve as fidelity repair (Sec.~\ref{sec:framework_routing}).

\tightparagraph{Observation 3: Approximate anchors seldom improve upon exact anchors.}
Only three of the 17 tested variants lie above the exact-anchor frontier
(Fig.~\ref{fig:characterization}(d); App.~\ref{sec:appendix_char_o3}).
Any approximation inside an anchor is inherited by every step until the next anchor.
\emph{Design consequence:} anchors should remain exact full computations (Sec.~\ref{sec:framework_schedule}).

\tightparagraph{Observation 4: Inter-anchor staleness is spatially heterogeneous.}
Between anchors, tokens go stale at different rates.
High-frequency reuse error grows faster in absolute terms than low-frequency (LF) error.
It remains about $2$--$2.5\times$ the LF error at cache ages 1--10, up to $2.7\times$ on
individual prompts (Fig.~\ref{fig:characterization_o4} in App.~\ref{sec:appendix_char_o4}).
Cache age complements drift: a token with small per-step drift can still become unreliable after many reuse steps.
\emph{Design consequence:} repair between anchors should follow measured staleness, proxied by frequency-resolved
drift and cache age
(Sec.~\ref{sec:design_risk}).

\tightparagraph{Summary:}
Because reactive signals report drift only after it occurs, these observations motivate the following control
principle: schedule exact synchronization in advance (Obs.~1, 3), while using measured drift only to allocate
inter-anchor repair (Obs.~2, 4).

\section{Frequency-Step Cache Design}
\label{sec:design}
\label{sec:method}
\Mname{} separates cache control into two decisions: exact synchronization determines \emph{when} the cache is
recomputed exactly, while adaptive repair determines \emph{how much} residual computation is spent between
synchronizations.
Residual caching replays a transformer block's stored residual instead of recomputing it.

\looseness=-1
After the warmup, \Mname{} recomputes all tokens in two situations that differ in timing.
An \emph{anchor} occurs at a step fixed before generation.
At non-anchor steps, the staleness risk $\rho_s$ (Sec.~\ref{sec:design_risk}) may select a \emph{full repair}.
The risk may trigger a full repair but never removes or delays an anchor.
\Mname{} maintains two states. The \emph{residual cache state} stores each token's whole-stack residual, while keys and
values are cached per block.
The \emph{anchor latent reference} $\mathbf{X}_{\mathrm{ref}}$ holds the latent bands stored at the last anchor
(Sec.~\ref{sec:design_risk}). All drift is measured against it.
An anchor replaces $\mathbf{X}_{\mathrm{ref}}$; a full repair keeps it.
Removing anchors and letting $\rho_s$ trigger every full synchronization after warmup lowers
fidelity (Sec.~\ref{sec:ablation_separation}).

\begin{figure*}[t]
\centering
\includegraphics[width=\linewidth]{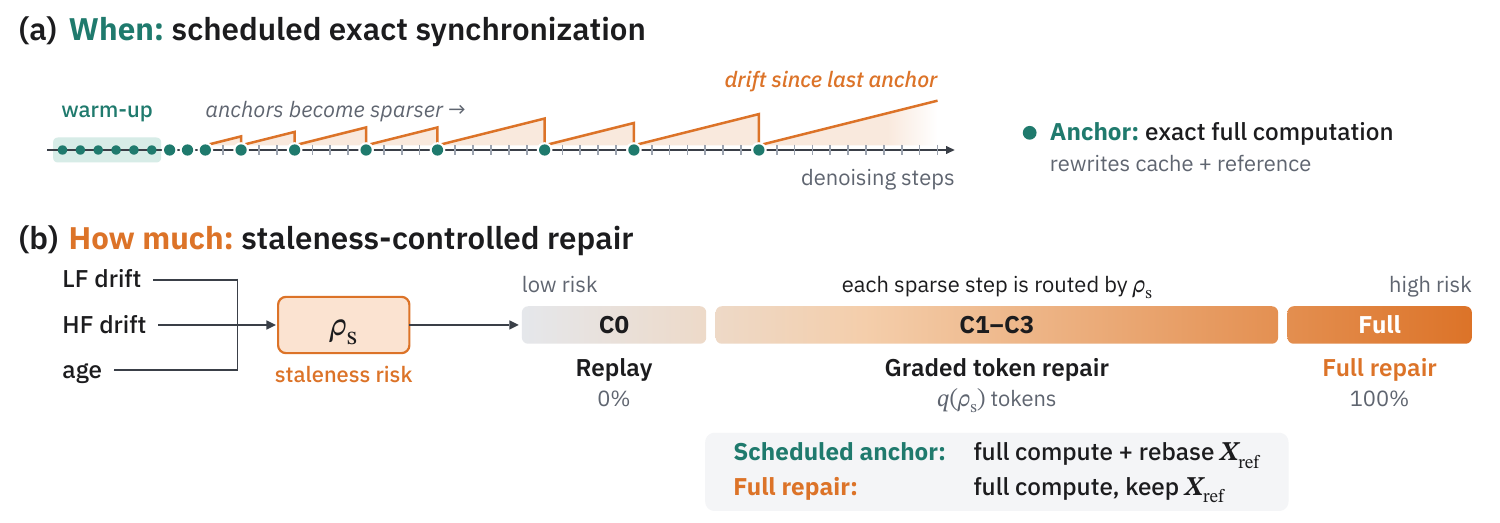}
\caption{\textbf{Anchors are fixed before generation; full repairs are chosen by the risk $\rho_s$ and leave the
schedule unchanged.}
\textbf{(a)} \emph{When}: dense warmup, then dense-to-sparse anchors.
\textbf{(b)} \emph{How much}: drift from $\mathbf{X}_{\mathrm{ref}}$ and cache age form $\rho_s$, which selects replay
($C0$), token repair ($C1$--$C3$), or full repair. An anchor replaces $\mathbf{X}_{\mathrm{ref}}$; a full repair keeps it.}
\label{fig:overview}
\end{figure*}

Each step makes up to three decisions (Fig.~\ref{fig:overview}; Algorithm~\ref{alg:fsc} in
App.~\ref{sec:appendix_design_algorithm}).
First, warmup steps and anchors compute all tokens exactly.
Second, at any other step, $\rho_s$ chooses how much to compute: replay ($C0$), token repair ($C1$--$C3$), or full repair.
Third, if token repair is selected, accumulated drift and token age choose which tokens to recompute.

\subsection{Dense-to-Sparse Anchor Schedule}
\label{sec:framework_schedule}
Because cacheability grows along the trajectory and approximate anchors seldom improve upon exact anchors
(Fig.~\ref{fig:characterization}(a,d)), \Mname{} places exact anchors densely early and sparsely late.
With warmup length $W$ (steps $s<W$ are fully computed), $n$ post-warmup anchors,
and last eligible step $E$, anchor $j$ is placed at
\begin{equation}
a_j = \operatorname{round}\!\left[\,W + c_j\,(E-W)\right],
\qquad
c_j = \frac{\sum_{i<j}(1+i)}{\sum_{i<n-1}(1+i)},
\qquad j=0,\ldots,n-1.
\label{eq:anchors}
\end{equation}
The nominal anchor gaps grow linearly ($1{:}2{:}3{:}\cdots$; rounding in App.~\ref{sec:appendix_design_schedule}).

\subsection{Frequency- and Age-Aware Staleness Estimation}
\label{sec:design_risk}
\begin{figure}[t]
\centering
\includegraphics[width=\linewidth]{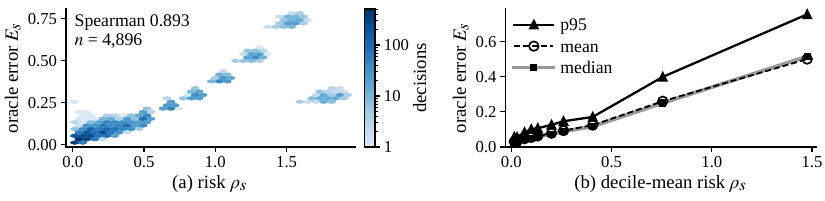}
\caption{\textbf{The staleness risk $\rho_s$ is strongly associated with exact reuse error (Spearman $0.893$) in an
offline diagnostic.}
SkyReels-V2 1.3B, $4{,}896$ non-anchor decisions on the eight development prompts (App.~\ref{sec:appendix_oracle}).
\textbf{(a)} $\rho_s$ versus the oracle reuse error $E_s$; within each prompt the Spearman correlation is
$0.909$--$0.972$.
\textbf{(b)} Mean, median, and $95$th-percentile $E_s$ per $\rho_s$ decile; mean and median rise monotonically.}
\label{fig:oracle_rho}
\vspace{-0.5em}
\end{figure}

Between anchors, staleness is spatially non-uniform; \Mname{} estimates it from the latent alone,
without evaluating the transformer. Let $\mathbf{X}_s$ be the latent at step $s$. Its per-frame
coarse and detail bands are
\begin{equation}
\mathbf{X}_{s,\mathrm{LF}}=\mathrm{Up}\!\left(\mathrm{AvgPool}_{2\times2}(\mathbf{X}_s)\right),
\qquad
\mathbf{X}_{s,\mathrm{HF}}=\mathbf{X}_s-\mathbf{X}_{s,\mathrm{LF}}.
\label{eq:bands}
\end{equation}
For token $i$ and band $\mathrm{B}\in\{\mathrm{LF},\mathrm{HF}\}$, drift from $\mathbf{X}_{\mathrm{ref}}$ is
\begin{equation}
d^{\mathrm{B}}_{s,i}
=\frac{\|\mathbf{X}_{s,\mathrm{B},i}-\mathbf{X}_{\mathrm{ref},\mathrm{B},i}\|_2}
      {\|\mathbf{X}_{\mathrm{ref},\mathrm{B},i}\|_2+\varepsilon}.
\label{eq:drift}
\end{equation}
Each token's age $a_{s,i}$ counts reuse steps since its last recompute.
The staleness risk $\rho_s$ is the step-level score that chooses the repair level.
It combines the $95$th percentile of each band's drift with the mean normalized age,
\begin{equation}
\rho_s = w_{\mathrm{HF}}\,Q_{0.95}\!\big(d^{\mathrm{HF}}_{s,\cdot}\big)
       + w_{\mathrm{LF}}\,Q_{0.95}\!\big(d^{\mathrm{LF}}_{s,\cdot}\big)
       + w_{\mathrm{age}}\,\frac{1}{N}\sum_{i=1}^{N}\frac{a_{s,i}}{k_{\mathrm{tok}}},
\label{eq:risk}
\end{equation}
where $N$ is the number of tokens and $k_{\mathrm{tok}}$ the maximum token age.
The risk excludes accumulated drift, which only selects tokens (Sec.~\ref{sec:framework_repair}).
We assess its association with the exact reuse error $E_s$, the $95$th-percentile relative error of cached residuals,
in an offline diagnostic on the eight development prompts (App.~\ref{sec:appendix_oracle}).
Over $4{,}896$ non-anchor decisions, $\rho_s$ reaches Spearman $0.893$ with $E_s$ (Fig.~\ref{fig:oracle_rho}).
As a detector of unsafe reuse, defined as $E_s>0.05$, it reaches ROC-AUC $0.930$ (Fig.~\ref{fig:oracle_validation}).

\subsection{Hierarchical Replay and Selective Repair}
\label{sec:framework_repair}
\label{sec:framework_routing}
\label{sec:framework_budget}
\label{sec:framework_state}
A non-anchor step is routed to one of five levels.
Replay ($C0$) reuses all cached residuals.
Token repair ($C1$--$C3$) recomputes increasing token fractions $q_m$: base fractions $\hat q_m$ times a common
scale $q_{\mathrm{scale}}$.
Full repair ($\mathrm{Full}$) recomputes all residuals.
The level rises as $\rho_s$ passes the step thresholds
\begin{equation}
t_m(s)=b_m\,\varphi_m(s)\,r,\qquad m\in\{\mathrm{skip},1,2,\mathrm{full}\},
\label{eq:thresholds}
\end{equation}
with base thresholds $b_m$, phase shaping $\varphi_m(s)$, and threshold scale $r$; $C0$ needs $\rho_s<t_{\mathrm{skip}}$.
Phase shaping tightens thresholds early and loosens them late.
A token is overdue once its age reaches $k_{\mathrm{tok}}$, and any overdue token forbids $C0$
(App.~\ref{sec:appendix_design_routing}).
Token repair ranks tokens by accumulated high-frequency drift ${A^{\mathrm{HF}}_{s,i}=\sum_{s'}d^{\mathrm{HF}}_{s',i}}$
since the last recompute of token $i$ (App.~\ref{sec:appendix_design_risk}).
The repaired tokens are the top fraction $q_m$ plus every overdue token:
\begin{equation}
\mathcal{S}_s=\mathrm{Top}_{q_m}\!\big(\{A^{\mathrm{HF}}_{s,i}\}_{i=1}^{N}\big)
\;\cup\;\{\,i:\;a_{s,i}\ge k_{\mathrm{tok}}\,\}.
\label{eq:repairset}
\end{equation}
At $C3$, an optional LF gate adds the top-$p_{\mathrm{LF}}$ tokens ranked by instantaneous low-frequency drift
$d^{\mathrm{LF}}_{s,i}$. The tokens in $\mathcal{S}_s$ run through every block as a gathered batch that attends to all cached keys and values
(App.~\ref{sec:appendix_design_state}).

All parameter values are listed in App.~\ref{sec:appendix_impl}.
On SkyReels-V2, \Mname{}-Speed differs from \Mname{}-Quality (Sec.~\ref{sec:exp_setup}) only in $r$, the
primary speed--quality control parameter, and in the warmup $W$.
On the full benchmark, \Mname{}-Quality routes $70.3\%$ / $23.8\%$ / $5.8\%$ of non-anchor
steps to $C0$ / $C1$--$C3$ / $\mathrm{Full}$ (\Mname{}-Speed: $69.1\%$ / $25.2\%$ / $5.7\%$).

\subsection{Backbone Instantiation and Overhead}
\label{sec:framework_backbones}
\label{sec:framework_complexity}
On MAGI-1, \Mname{} scales the native token-level reuse threshold, forces exact computation at anchors, and
fully recomputes every $k_{\mathrm{tok}}$-th step (App.~\ref{sec:appendix_design_backbones}).
MAGI-1 therefore tests transfer through its native cache interface, not the frequency controller.
The controller needs no auxiliary network and adds under $0.2\%$ latency (App.~\ref{sec:appendix_design_complexity}).

\section{Experiments}
\label{sec:experiments}

\subsection{Experimental Setup}
\label{sec:exp_setup}

\paragraph{Base Models and Baselines.}
SkyReels-V2, FramePack~\citep{zhang2025framepack}, and Matrix-Game-3.0~\citep{wang2026matrixgame3} use the full
\Mname{} controller, whereas MAGI-1 tests transfer through its native cache interface (Sec.~\ref{sec:framework_backbones}).
SkyReels-V2-1.3B~\citep{chen2025skyreelsv2} generates 540P videos of $177$ frames (two $97$-frame chunks with $17$
overlapping frames, $50$ steps per temporal group; App.~\ref{sec:appendix_impl}).
MAGI-1-4.5B-distill~\citep{teng2025magi1} generates $720\times720$ videos of $168$ frames (seven $24$-frame chunks,
$64$ steps per chunk).
We compare against vanilla, FlowCache~\citep{ma2026flowcache} with chunk-level reuse,
MotionCache~\citep{xu2026motioncache} with token-level reuse in its released configuration, and our
latency-calibrated SeaCache port.
All share one stack, prompts, seed, checkpoints, and settings; SCOPE~\citep{scope2026} runs on its official stack.
FlowCache runs with its KV-cache compression disabled.
In a probe run, the official FlowCache stack without compression runs at $2.21\times$, close to our
reproduction ($2.04\times$), at comparable fidelity (App.~\ref{sec:appendix_search}).
We found no public implementation of ARCache~\citep{nan2026arcache}.

\begin{table}[t]
\centering
\caption{
\textbf{\Mname{}-Quality matches MotionCache's speed with higher fidelity, and \Mname{}-Speed is faster while keeping
higher SSIM and PSNR.}
Speed and PFLOPs: Bench (MAGI-1: speedups on the vanilla-clean prompt pool after per-method latency auditing, MotionCache and \Mname{} PFLOPs on a
$20$-prompt sample); VBench: the $505$ Bench prompts of the evaluated dimensions ($500$ on MAGI-1); PSNR/SSIM/LPIPS: Fid ($110$ on MAGI-1).
PFLOPs come from executed-token counters, warmup forwards included, on a different basis per backbone;
Mem: peak memory (both App.~\ref{sec:appendix_impl}).
$^{\ast}$Our latency-calibrated SeaCache port (App.~\ref{sec:appendix_matched}).
$^{\dagger}$\mbox{MAGI-1} tests transfer through its native cache interface (Sec.~\ref{sec:framework_backbones}).
Best accelerated result per backbone in \textbf{bold}, second best \underline{underlined}.
}
\label{tab:main_results}

\small
\setlength{\tabcolsep}{3pt}

\begin{tabular}{@{}llcccccccc@{}}
\toprule
Model & Method & PFLOPs & Speedup & Latency & Mem & VBench & PSNR & SSIM & LPIPS \\
 & & $\downarrow$ & $\uparrow$ & (s)~$\downarrow$ & (GB)~$\downarrow$ & $\uparrow$ & $\uparrow$ & $\uparrow$ & $\downarrow$ \\
\midrule
\multirow{6}{*}{SkyReels-V2}
& Vanilla      & 37.7 & 1.000$\times$ & 367 & 53.7 & 0.7977 & -- & -- & -- \\
& FlowCache    & 17.4 & 2.037$\times$ & 180 & \textbf{58.1} & \textbf{0.7939} & 17.57 & 0.7946 & \textbf{0.3073} \\
& MotionCache  & 15.4 & 2.275$\times$ & 161 & 58.9 & 0.7898 & 17.34 & 0.7899 & 0.3366 \\
& SeaCache-AR$^{\ast}$ & 14.9 & \underline{2.320$\times$} & \underline{158} & 58.6 & 0.7886 & 17.52 & 0.7919 & 0.3331 \\
& \Mname{}-Quality (ours) & \underline{14.8} & 2.275$\times$ & 161 & \underline{58.4} & \underline{0.7900} & \textbf{18.38} & \textbf{0.8209} & \underline{0.3122} \\
& \Mname{}-Speed (ours)   & \textbf{14.1} & \textbf{2.356$\times$} & \textbf{156} & \underline{58.4} & 0.7892 & \underline{17.87} & \underline{0.8088} & 0.3313 \\
\midrule
\multirow{5}{*}{MAGI-1}
& Vanilla      & 140.6 & 1.000$\times$ & 684 & 31.8 & 0.7969 & -- & -- & -- \\
& FlowCache    & \textbf{82.5} & \textbf{1.486$\times$} & \textbf{461} & 19.7 & \textbf{0.7991} & 16.64 & 0.7782 & 0.3559 \\
& MotionCache  & 90.4 & 1.357$\times$ & 504 & 19.7 & 0.7950 & 18.76 & 0.8468 & 0.2778 \\
& \Mname{}-Quality (ours)$^{\dagger}$ & 86.8 & 1.421$\times$ & 482 & 19.7 & \underline{0.7954} & \textbf{20.49} & \textbf{0.8883} & \textbf{0.2297} \\
& \Mname{}-Speed (ours)$^{\dagger}$ & \underline{84.6} & \underline{1.457$\times$} & \underline{470} & 19.7 & 0.7896 & \underline{20.00} & \underline{0.8782} & \underline{0.2460} \\
\bottomrule
\end{tabular}
\vspace{-0.5em}
\end{table}

\paragraph{Operating Points.}
We report two operating points of the same controller. On SkyReels-V2, \Mname{}-Quality matches MotionCache's
latency and is the single configuration used for all SkyReels-V2 ablations, sensitivity and seed studies, and the 14B
no-retuning transfer; \Mname{}-Speed differs only in warmup length and threshold scale and is a faster point on
the speed--quality frontier (App.~\ref{sec:appendix_pareto}). On MAGI-1, both points are calibrated on its native cache
interface and run faster than MotionCache. FramePack and Matrix-Game each use one per-backbone calibrated
configuration (App.~\ref{sec:appendix_crossfamily}).

\paragraph{Evaluation Sets and Metrics.}
Quality is measured with VBench-Long~\citep{huang2024vbench,huang2025vbenchpp} on a fixed dimension set chosen with
reference to prior autoregressive video caching protocols~\citep{ma2026flowcache,xu2026motioncache}, with the official
normalization and weights (App.~\ref{sec:appendix_impl}).
Unless noted, all timings use one NVIDIA RTX~PRO 6000 GPU per video on otherwise idle devices.
Fidelity is PSNR, SSIM, and LPIPS~\citep{lpips} against the same-seed vanilla output.
VBench and same-seed fidelity are complementary: VBench measures reference-free video quality, whereas fidelity measures
the approximation error that caching introduces relative to the unmodified generator.
Same-seed fidelity requires vanilla references and is therefore evaluated on subsets. The evaluation sets are:
\par\vspace{0.4em}
{\centering
\small
\setlength{\tabcolsep}{4.5pt}
\begin{tabular}{@{}lrll@{}}
\toprule
Set & Prompts & Selection & Used for \\
\midrule
Bench      & $946$ & full VBench-Long benchmark               & speed, compute, VBench \\
Fid        & $102$ & first ten of each prompt set, minus Dev  & fidelity in Table~\ref{tab:main_results} (MAGI-1: all $110$) \\
Paired     & $235$ & stratified subset, minus Dev             & equal latency (Table~\ref{tab:transfer}(a), Fig.~\ref{fig:mc_frontier}(a)) \\
Dev        & $8$   & our development set (App.~\ref{sec:appendix_impl}) & ablations, sensitivity, sweep, Fig.~\ref{fig:oracle_rho} \\
Cross      & $40$  & per backbone, never used for calibration & cross-family separation test (Sec.~\ref{sec:ablation_separation}) \\
14B        & $47$  & five per prompt set, minus Dev           & SkyReels-V2 14B (Table~\ref{tab:transfer}(b)) \\
\bottomrule
\end{tabular}
\par}

\subsection{Main Results}
\label{sec:main_results}

\paragraph{Higher fidelity at equal latency.}
On SkyReels-V2 (Table~\ref{tab:main_results}; \Mname{}-Speed examples in Fig.~\ref{fig:qualitative}), \Mname{}-Quality
matches MotionCache's latency. It raises SSIM from $0.790$ to $0.821$, improves PSNR by $1.04$~dB, and lowers LPIPS by
$0.024$; all three paired 95\% CIs exclude zero. \Mname{}-Speed reaches $2.36\times$, exceeding all shared-stack
baselines while improving SSIM by $0.019$ and PSNR by $0.53$~dB over MotionCache, both with CIs excluding zero. At both operating points the VBench difference from MotionCache is not statistically
resolved. This complements the fidelity result: the methods retain similar reference-free quality but differ in how
strongly caching perturbs the same-seed trajectory.

\begin{figure}[t]
\centering
\includegraphics[width=\linewidth]{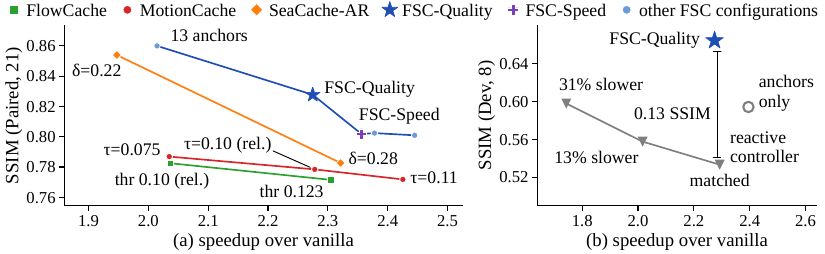}
\caption{\textbf{Operating-point robustness and the effect of separating synchronization from repair.}
\textbf{(a)} SkyReels-V2 1.3B; speedup: mean over the $235$ Paired prompts; SSIM: their $21$ prompts with vanilla references. Outside the MotionCache threshold sweep, \Mname{} retains higher SSIM than the shared-stack
baselines at comparable speed, and MotionCache's released $\tau{=}0.10$ is its latency-matched point. Lines join each
method's operating points in speed order ($\tau$, thr, $\delta$: MotionCache, FlowCache, and SeaCache-AR thresholds; rel.: released;
\Mname{}: \Mname{}-Quality with $13$ anchors and $r{=}1.40$, \Mname{}-Quality, \Mname{}-Speed, and two other $W{=}5$ variants
of Table~\ref{tab:full_bench_all}).
\textbf{(b)} Dev set (Table~\ref{tab:ablation}); speedup: mean over its $8$ prompts. The purely reactive variant remains
below \Mname{}-Quality at every tested latency; its SSIM gap narrows from $0.13$ (matched latency) to $0.07$ ($31\%$ more latency); hollow: anchors only,
no repair.}
\label{fig:mc_frontier}
\vspace{-1em}
\end{figure}

\paragraph{Baseline operating points.}
The MotionCache threshold sweep confirms that its released $\tau{=}0.10$ is also its latency-matched operating point.
On evaluation prompts outside the sweep set, \Mname{}-Quality retains a $+0.049$ SSIM advantage at matched latency.
At the same operating point, \Mname{}-Quality also exceeds matched FlowCache and SeaCache-AR by $0.056$ and $0.045$ SSIM,
respectively (Table~\ref{tab:transfer}(a), Fig.~\ref{fig:mc_frontier}(a)).
The concurrent SCOPE is faster in seconds but $0.015$ lower on VBench than \Mname{}-Quality (CI $[-0.024,-0.006]$);
stack and fidelity comparability are detailed in App.~\ref{sec:appendix_scope}.

\paragraph{No-retuning transfer to 14B.}
\Mname{}-Quality transfers unchanged to the ${\sim}11\times$ larger SkyReels-V2 14B, reaching $2.12\times$
versus MotionCache's $1.79\times$ with higher SSIM and PSNR, slightly higher LPIPS, and comparable VBench quality
(Table~\ref{tab:transfer}(b); App.~\ref{sec:appendix_14b}).
On 1.3B, the fidelity gain replicates across five seeds (App.~\ref{sec:appendix_seeds}).
The advantage also grows with motion: across vanilla motion terciles, \Mname{}-Quality's SSIM margin over MotionCache
increases from $+0.018$ at low motion to $+0.049$ at high motion, with the same trend in LPIPS.

\paragraph{Transfer to MAGI-1.}
Both \Mname{} operating points improve speed and same-seed fidelity over MotionCache at comparable VBench quality:
\Mname{}-Quality raises SSIM from $0.847$ to $0.888$, and \Mname{}-Speed reaches $0.878$ at higher speed
(Table~\ref{tab:main_results}). FlowCache is fastest but has substantially lower same-seed fidelity; motion-dependent
behavior is analyzed in App.~\ref{sec:appendix_failure}.

\begin{table}[t]
\centering
\caption{\textbf{\Mname{} maintains higher fidelity at matched latency and higher SSIM and PSNR on 14B without retuning.}
Paired: speedup on all $235$ prompts, VBench on $126$, fidelity on $21$. 14B ($352\times640$): VBench is evaluated on
$25$ of the $47$ prompts; all other metrics use all $47$. Best in \textbf{bold}, second best \underline{underlined} per block.}
\label{tab:transfer}
\vspace{2pt}
\small
\begin{tabular}{@{}lccccc@{}}
\toprule
Method & Speedup$\uparrow$ & VBench$\uparrow$ & PSNR$\uparrow$ & SSIM$\uparrow$ & LPIPS$\downarrow$ \\
\midrule
\multicolumn{6}{@{}l}{\emph{(a) Equal latency, SkyReels-V2 1.3B (Paired)}} \\
MotionCache & 2.274$\times$ & 0.7885 & 17.55 & 0.7789 & 0.3156 \\
FlowCache (threshold $0.123$) & \underline{2.305$\times$} & \textbf{0.7922} & 17.46 & 0.7715 & \underline{0.3125} \\
SeaCache-AR (threshold $\delta{=}0.28$) & \textbf{2.321$\times$} & \underline{0.7908} & \underline{17.68} & \underline{0.7826} & 0.3222 \\
\Mname{}-Quality & 2.275$\times$ & 0.7883 & \textbf{18.87} & \textbf{0.8277} & \textbf{0.2840} \\
\midrule
\multicolumn{6}{@{}l}{\emph{(b) SkyReels-V2 14B, 1.3B configuration unchanged (14B)}} \\
FlowCache & 1.595$\times$ & 0.8292 & \underline{16.07} & \underline{0.7692} & \textbf{0.3189} \\
MotionCache & \underline{1.792$\times$} & \underline{0.8327} & 15.88 & 0.7600 & \underline{0.3289} \\
\Mname{}-Quality & \textbf{2.121$\times$} & \textbf{0.8406} & \textbf{16.18} & \textbf{0.7745} & 0.3302 \\
\bottomrule
\end{tabular}
\vspace{-0.5em}
\end{table}

\subsection{Ablation Studies}
\label{sec:ablation}

\begin{table}[t]
\centering
\caption{
\textbf{Separating synchronization from repair: a purely reactive variant reduces SSIM by about $0.13$ at matched
latency; increasing its latency by $13\%$ or $31\%$ narrows but does not close the gap.}
SkyReels-V2, Dev set; rows modify \Mname{}-Quality (D2S: dense-to-sparse; purely reactive: no anchors, $\rho_s$ triggers
repair and full synchronization; $r$: threshold scale); s/video: median wall-clock. Absolute scores are not comparable with Table~\ref{tab:main_results} (different prompts and fidelity harness). The last two columns count the prompts on which
\Mname{}-Quality outperforms each variant; further variants in App.~\ref{sec:appendix_separation}. Best in \textbf{bold}, second best \underline{underlined}.
}
\label{tab:ablation}

\small
\setlength{\tabcolsep}{3pt}

\begin{tabular}{@{}lccccc@{}}
\toprule
Variant & SSIM$\uparrow$ & LPIPS$\downarrow$ & s/video$\downarrow$ &
\multicolumn{2}{c}{Wins (of 8)} \\
\cmidrule(lr){5-6}
& & & & SSIM & LPIPS \\
\midrule
\Mname{}-Quality: D2S anchors + repair & \textbf{0.6641} & \textbf{0.2925} & 161 & -- & -- \\
\midrule
purely reactive, matched latency ($r{=}1.076$) & 0.5332 & 0.4786 & 160 & 8 & 8 \\
purely reactive, $13\%$ slower ($r{=}0.631$) & 0.5575 & 0.4285 & 182 & 8 & 8 \\
purely reactive, $31\%$ slower ($r{=}0.250$) & 0.5976 & 0.3569 & 211 & 8 & 6 \\
anchors only, no repair ($13$ D2S anchors) & 0.5940 & 0.3337 & \textbf{153} & 8 & 8 \\
\midrule
uniform anchor spacing (with repair) & \underline{0.6441} & \underline{0.3189} & 157 & 5 & 5 \\
sparse-to-dense, reversed (with repair) & 0.6033 & 0.3659 & 161 & 8 & 8 \\
no age term, no overdue safeguard & 0.6429 & 0.3369 & \underline{155} & 8 & 8 \\
\bottomrule
\end{tabular}
\vspace{-1em}
\end{table}

\paragraph{Separating Synchronization from Repair.}
\label{sec:ablation_separation}
Removing the anchors and letting $\rho_s$ trigger both repair and full synchronization gives a controlled purely
reactive variant of \Mname{}: it uses the same risk signal and repair machinery but has no scheduled anchors.
At matched latency, \Mname{}-Quality improves SSIM by $0.131$ and lowers LPIPS by $0.186$ on all eight development
prompts. Giving the reactive variant $13\%$ or $31\%$ more latency narrows but does not eliminate the SSIM gap
(Fig.~\ref{fig:mc_frontier}(b); Table~\ref{tab:ablation}).
This controlled result shows that, for \Mname{}'s risk signal and repair mechanism, keeping exact synchronization
scheduled is more effective than using the same reactive signal to determine both decisions.
On the Cross set, \Mname{} also outperforms a latency-matched purely reactive comparator on FramePack / Matrix-Game: SSIM
$+0.079$ / $+0.058$, PSNR $+5.18$ / $+2.77$~dB, and LPIPS $-0.092$ / $-0.057$; all 95\% CIs exclude zero.
These backbones differ from SkyReels-V2 in cache interface and token-execution cost, so the result supports the
separation principle independently of SkyReels-specific sparse execution.

\paragraph{Schedule, Age, and Sensitivity.}
\label{sec:ablation_schedule}
\label{sec:ablation_controller}
A $2\times2$ factorial on the Dev set separates the two components: dense-to-sparse placement adds $0.017$ SSIM without
repair, adaptive repair adds $0.068$ under uniform anchors, while combining both yields the largest gain of $0.088$.
Two results support the growing gaps of Eq.~\ref{eq:anchors}: the influence of a
single cached step on the final video decays $6.1\times$ from step $8$ to step $39$,
and every alternative schedule evaluated end to end, including uniform, reversed, and three surrogate-optimized
schedules, yields lower mean SSIM (Tables~\ref{tab:ablation} and~\ref{tab:anchor_search}).
Cache age provides an important overdue safeguard: removing the age term and
its overdue rule reduces SSIM by $0.021$ on all eight prompts while lowering latency
by $4\%$. Around \Mname{}-Quality, six of eight local perturbations change SSIM by $\le0.003$ and latency by
$\le1.8\%$.

\section{Conclusion}
\label{sec:conclusion}

Our results support one principle for cache control in AR video diffusion: for reactive staleness
signals, exact synchronization should be scheduled before measured drift accumulates, and measured
drift should only decide how much repair to spend between synchronizations.
\Mname{} instantiates this principle through scheduled exact anchors and whole-step replay,
with a frequency- and age-aware latent risk controlling inter-anchor repair.
On SkyReels-V2 (1.3B and, unchanged, 14B), \Mname{} is closer to the
vanilla trajectory than MotionCache in SSIM and PSNR at equal or higher speed, and coupling both decisions into one risk signal reduces fidelity
at matched latency (Sec.~\ref{sec:ablation}).
On MAGI-1, \Mname{}'s native-interface port improves speed and fidelity over MotionCache at comparable VBench, and
the separation advantage holds on FramePack and Matrix-Game with per-backbone calibration.

\section*{AI use statement}
We used generative AI tools, primarily Claude by Anthropic, as coding and writing assistants. Their
uses included implementing author-specified methods and integrations, writing baseline ports and
experiment scripts, suggesting ablations and experimental parameters, assisting with experiment
planning, drafting and editing text, translating author notes into English, generating code, auditing
reported numbers against experiment logs, and finding and summarizing related work.

Generative AI also assisted in brainstorming hypotheses, interpreting experimental outputs, and
refining the conceptual presentation of the work. All research questions, methodological choices,
experimental protocols, acceptance or rejection of suggested analyses, and final scientific claims
were determined by the authors. We did not use generative AI to generate synthetic datasets.

All AI-assisted outputs were reviewed by the authors. AI-assisted code was validated through
consistency tests and reproduction of the reported experiments. Reported numerical results were
independently re-derived from the underlying experiment databases and checked by the authors, and
descriptions of related work were checked against the cited papers. The authors take full
responsibility for the final text, methods, code, analyses, and conclusions.

\section*{Reproducibility Statement}
Sec.~\ref{sec:method} defines the \Mname{} controller, including the dense-to-sparse anchor schedule, the frequency-
and age-aware staleness risk, and the repair routing. Algorithm~\ref{alg:fsc} (App.~\ref{sec:appendix_design_algorithm}) gives the per-step
procedure, while App.~\ref{sec:appendix_design} provides the exact schedule construction, routing rules, cache-state updates,
and backbone-specific instantiations (App.~\ref{sec:appendix_design_backbones}).

App.~\ref{sec:appendix_impl} reports the hardware and software environment, generation settings and seeds, timing,
FLOPs, and memory measurement procedures, and the final hyperparameters for each backbone.
App.~\ref{sec:appendix_search} describes baseline configuration and reproduction and reports the evaluated final operating
points (Tables~\ref{tab:full_bench_all} and~\ref{tab:magi_points}). The evaluation protocol is specified in Sec.~\ref{sec:exp_setup}.

\bibliographystyle{plainnat}
\bibliography{references}

@misc{chen2025skyreelsv2, 
  title={{SkyReels-V2}: Infinite-length Film Generative Model},
  author={Chen, Guibin and Lin, Dixuan and Yang, Jiangping and Lin, Chunze and Zhu, Junchen and Fan, Mingyuan and Zhang, Hao and Chen, Sheng and Chen, Zheng and Ma, Chengcheng and Xiong, Weiming and Wang, Wei and Pang, Nuo and Kang, Kang and Xu, Zhiheng and Jin, Yuzhe and Liang, Yupeng and Song, Yubing and Zhao, Peng and Xu, Boyuan and Qiu, Di and Li, Debang and Fei, Zhengcong and Li, Yang and Zhou, Yahui},
  year={2025},
  eprint={2504.13074},
  archivePrefix={arXiv},
  primaryClass={cs.CV},
  url={https://arxiv.org/abs/2504.13074}
}

@inproceedings{xu2026motioncache,
  title={Motion-Aware Caching for Efficient Autoregressive Video Generation},
  author={Xu, Jing and Ma, Yuexiao and Zheng, Xuzhe and Wang, Xing and Liu, Shiwei and Yan, Chenqian and Zheng, Xiawu and Ji, Rongrong and Chao, Fei and Liu, Songwei},
  booktitle={Proceedings of the 43rd International Conference on Machine Learning},
  year={2026}
}

@inproceedings{ma2026flowcache,
  title={Flow Caching for Autoregressive Video Generation},
  author={Ma, Yuexiao and Zheng, Xuzhe and Xu, Jing and Xu, Xiwei and Ling, Feng and Zheng, Xiawu and Kuang, Huafeng and Li, Huixia and Wang, Xing and Xiao, Xuefeng and Chao, Fei and Ji, Rongrong},
  booktitle={International Conference on Learning Representations},
  year={2026}
}

@misc{selvaraju2024fora,
  title={{FORA}: Fast-Forward Caching in Diffusion Transformer Acceleration},
  author={Selvaraju, Pratheba and Ding, Tianyu and Chen, Tianyi and Zharkov, Ilya and Liang, Luming},
  year={2024},
  eprint={2407.01425},
  archivePrefix={arXiv},
  primaryClass={cs.CV},
  url={https://arxiv.org/abs/2407.01425}
}

@misc{chen2024deltadit,
  title={{$\Delta$-DiT}: A Training-Free Acceleration Method Tailored for Diffusion Transformers},
  author={Chen, Pengtao and Shen, Mingzhu and Ye, Peng and Cao, Jianjian and Tu, Chongjun and Bouganis, Christos-Savvas and Zhao, Yiren and Chen, Tao},
  year={2024},
  eprint={2406.01125},
  archivePrefix={arXiv},
  primaryClass={cs.CV},
  url={https://arxiv.org/abs/2406.01125}
}

@inproceedings{liu2024teacache,
  title={Timestep Embedding Tells: It's Time to Cache for Video Diffusion Model},
  author={Liu, Feng and Zhang, Shiwei and Wang, Xiaofeng and Wei, Yujie and Qiu, Haonan and Zhao, Yuzhong and Zhang, Yingya and Ye, Qixiang and Wan, Fang},
  booktitle={Proceedings of the IEEE/CVF Conference on Computer Vision and Pattern Recognition},
  pages={7353--7363},
  year={2025},
  url={https://openaccess.thecvf.com/content/CVPR2025/html/Liu_Timestep_Embedding_Tells_Its_Time_to_Cache_for_Video_Diffusion_CVPR_2025_paper.html}
}

@inproceedings{bu2025dicache,
  title={{DiCache}: Let Diffusion Model Determine Its Own Cache},
  author={Bu, Jiazi and Ling, Pengyang and Zhou, Yujie and Wang, Yibin and Zang, Yuhang and Lin, Dahua and Wang, Jiaqi},
  booktitle={International Conference on Learning Representations},
  year={2026}
}

@inproceedings{zou2025toca,
  title={Accelerating Diffusion Transformers with Token-wise Feature Caching},
  author={Zou, Chang and Liu, Xuyang and Liu, Ting and Huang, Siteng and Zhang, Linfeng},
  booktitle={International Conference on Learning Representations},
  year={2025},
  eprint={2410.05317},
  archivePrefix={arXiv},
  url={https://arxiv.org/abs/2410.05317}
}

@misc{liu2025freqca,
  title={{FreqCa}: Accelerating Diffusion Models via Frequency-Aware Caching},
  author={Liu, Jiacheng and Cai, Peiliang and Zhou, Qinming and Lin, Yuqi and Kong, Deyang and Huang, Benhao and Pan, Yupei and Xu, Haowen and Zou, Chang and Tang, Junshu and Zheng, Shikang and Zhang, Linfeng},
  year={2025},
  eprint={2510.08669},
  archivePrefix={arXiv},
  primaryClass={cs.CV},
  url={https://arxiv.org/abs/2510.08669}
}

@misc{li2026spectralcache,
  title={Frequency-Aware Error-Bounded Caching for Accelerating Diffusion Transformers},
  author={Li, Guandong},
  year={2026},
  eprint={2603.05315},
  archivePrefix={arXiv},
  primaryClass={cs.CV},
  url={https://arxiv.org/abs/2603.05315}
}

@inproceedings{nan2026arcache,
  title={Accelerating Autoregressive Video Diffusion via History-Guided Cache and Residual Correction},
  author={Nan, Kepan and Zhao, Wangbo and Zhou, Penghao and Li, Jun and Yang, Zhenheng and Yang, Jian and Tai, Ying},
  booktitle={Proceedings of the IEEE/CVF Conference on Computer Vision and Pattern Recognition},
  year={2026},
  pages={43740--43750},
  url={https://openaccess.thecvf.com/content/CVPR2026/html/Nan_Accelerating_Autoregressive_Video_Diffusion_via_History-Guided_Cache_and_Residual_Correction_CVPR_2026_paper.html}
}

@inproceedings{huang2024vbench,
  title     = {{VBench}: Comprehensive Benchmark Suite for Video Generative Models},
  author    = {Huang, Ziqi and He, Yinan and Yu, Jiashuo and Zhang, Fan and
               Si, Chenyang and Jiang, Yuming and Zhang, Yuanhan and Wu, Tianxing and
               Jin, Qingyang and Chanpaisit, Nattapol and Wang, Yaohui and
               Chen, Xinyuan and Wang, Limin and Lin, Dahua and Qiao, Yu and Liu, Ziwei},
  booktitle = {Proceedings of the IEEE/CVF Conference on Computer Vision and Pattern Recognition},
  pages     = {21807--21818},
  year      = {2024}
}

@inproceedings{liu2025taylorseer,
  title={From Reusing to Forecasting: Accelerating Diffusion Models with {TaylorSeers}},
  author={Liu, Jiacheng and Zou, Chang and Lyu, Yuanhuiyi and Chen, Junjie and Zhang, Linfeng},
  booktitle={Proceedings of the IEEE/CVF International Conference on Computer Vision},
  pages={15853--15863},
  year={2025},
  eprint={2503.06923},
  archivePrefix={arXiv},
  primaryClass={cs.CV},
  url={https://openaccess.thecvf.com/content/ICCV2025/html/Liu_From_Reusing_to_Forecasting_Accelerating_Diffusion_Models_with_TaylorSeers_ICCV_2025_paper.html}
}

@inproceedings{chung2026seacache,
  title={{SeaCache}: Spectral-Evolution-Aware Cache for Accelerating Diffusion Models},
  author={Chung, Jiwoo and Hyun, Sangeek and Lee, MinKyu and Han, Byeongju and Cha, Geonho and Wee, Dongyoon and Hong, Youngjun and Heo, Jae-Pil},
  booktitle={Proceedings of the IEEE/CVF Conference on Computer Vision and Pattern Recognition},
  pages={14283--14294},
  year={2026}
}

@article{huang2025vbenchpp,
  title={{VBench++}: Comprehensive and Versatile Benchmark Suite for Video Generative Models},
  author={Huang, Ziqi and Zhang, Fan and Xu, Xiaojie and He, Yinan and Yu, Jiashuo and Dong, Ziyue and Ma, Qianli and Chanpaisit, Nattapol and Si, Chenyang and Jiang, Yuming and Wang, Yaohui and Chen, Xinyuan and Chen, Ying-Cong and Wang, Limin and Lin, Dahua and Qiao, Yu and Liu, Ziwei},
  journal={IEEE Transactions on Pattern Analysis and Machine Intelligence},
  year={2025},
  doi={10.1109/TPAMI.2025.3633890}
}

@inproceedings{lpips,
  title={The Unreasonable Effectiveness of Deep Features as a Perceptual Metric},
  author={Zhang, Richard and Isola, Phillip and Efros, Alexei A. and Shechtman, Eli and Wang, Oliver},
  booktitle={Proceedings of the IEEE Conference on Computer Vision and Pattern Recognition},
  year={2018},
  pages={586--595}
}

@inproceedings{yin2025causvid,
  title={From Slow Bidirectional to Fast Autoregressive Video Diffusion Models},
  author={Yin, Tianwei and Zhang, Qiang and Zhang, Richard and Freeman, William T. and Durand, Fr{\'e}do and Shechtman, Eli and Huang, Xun},
  booktitle={Proceedings of the IEEE/CVF Conference on Computer Vision and Pattern Recognition},
  pages={22963--22974},
  year={2025}
}

@inproceedings{peebles2023dit,
  title={Scalable Diffusion Models with Transformers},
  author={Peebles, William and Xie, Saining},
  booktitle={Proceedings of the IEEE/CVF International Conference on Computer Vision},
  pages={4195--4205},
  year={2023}
}

@misc{scope2026,
  title         = {Not All Frames Deserve Full Computation: Accelerating Autoregressive
                   Video Generation via Selective Computation and Predictive Extrapolation},
  author        = {Cui, Hanshuai and Tang, Zhiqing and Yao, Zhi and Meng, Fanshuai and
                   Jia, Weijia and Zhao, Wei},
  year          = {2026},
  eprint        = {2604.02979},
  note          = {To appear at ACM Multimedia 2026},
  archivePrefix = {arXiv},
  primaryClass  = {cs.CV},
  url           = {https://arxiv.org/abs/2604.02979}
}

@inproceedings{ma2024deepcache,
  title={{DeepCache}: Accelerating Diffusion Models for Free},
  author={Ma, Xinyin and Fang, Gongfan and Wang, Xinchao},
  booktitle={Proceedings of the IEEE/CVF Conference on Computer Vision and Pattern Recognition},
  pages={15762--15772},
  year={2024}
}

@inproceedings{peng2026ertacache,
  title={{ERTACache}: Error Rectification and Timesteps Adjustment for Efficient Diffusion},
  author={Peng, Xurui and Yan, Chenqian and Liu, Hong and Ma, Rui and Chen, Fangmin and Wang, Xing and Wu, Zhihua and Liu, Songwei and Lin, Mingbao},
  booktitle={International Conference on Learning Representations},
  year={2026}
}

@inproceedings{chen2024diffusionforcing,
  title={Diffusion Forcing: Next-token Prediction Meets Full-Sequence Diffusion},
  author={Chen, Boyuan and Mart{\'\i} Mons{\'o}, Diego and Du, Yilun and Simchowitz, Max and Tedrake, Russ and Sitzmann, Vincent},
  booktitle={Advances in Neural Information Processing Systems},
  volume={37},
  year={2024},
  url={https://proceedings.neurips.cc/paper_files/paper/2024/hash/2aee1c4159e48407d68fe16ae8e6e49e-Abstract-Conference.html}
}

@misc{teng2025magi1,
  title={{MAGI-1}: Autoregressive Video Generation at Scale},
  author={{Sand.ai} and Teng, Hansi and Jia, Hongyu and Sun, Lei and Li, Lingzhi and Li, Maolin and Tang, Mingqiu and Han, Shuai and Zhang, Tianning and Zhang, W. Q. and Luo, Weifeng and Kang, Xiaoyang and Sun, Yuchen and Cao, Yue and Huang, Yunpeng and Lin, Yutong and Fang, Yuxin and Tao, Zewei and Zhang, Zheng and Wang, Zhongshu and Liu, Zixun and Shi, Dai and Su, Guoli and Sun, Hanwen and Pan, Hong and Wang, Jie and Sheng, Jiexin and Cui, Min and Hu, Min and Yan, Ming and Yin, Shucheng and Zhang, Siran and Liu, Tingting and Yin, Xianping and Yang, Xiaoyu and Song, Xin and Hu, Xuan and Zhang, Yankai and Li, Yuqiao},
  year={2025},
  eprint={2505.13211},
  archivePrefix={arXiv},
  primaryClass={cs.CV},
  url={https://arxiv.org/abs/2505.13211}
}

@inproceedings{cui2026dpcache,
  title={Denoising as Path Planning: Training-Free Acceleration of Diffusion Models with {DPCache}},
  author={Cui, Bowen and Wang, Yuanbin and Xu, Huajiang and Chen, Biaolong and Zhang, Aixi and Jiang, Hao and Jin, Zhengzheng and Liu, Xu and Huang, Pipei},
  booktitle={Proceedings of the IEEE/CVF Conference on Computer Vision and Pattern Recognition},
  year={2026},
  pages={43632--43642},
  url={https://openaccess.thecvf.com/content/CVPR2026/html/Cui_Denoising_as_Path_Planning_Training-Free_Acceleration_of_Diffusion_Models_with_CVPR_2026_paper.html}
}

@article{zou2024duca,
  title         = {Rethinking Token-wise Feature Caching: Accelerating Diffusion Transformers
                   with Dual Feature Caching},
  author        = {Zou, Chang and Zheng, Shikang and Zhang, Evelyn and Guo, Runlin and
                   Xu, Haohang and Shi, Zhengyi and He, Conghui and Hu, Xuming and Zhang, Linfeng},
  journal       = {IEEE Transactions on Image Processing},
  volume        = {35},
  pages         = {6211--6220},
  year          = {2026},
  doi           = {10.1109/TIP.2026.3698363},
  eprint        = {2412.18911},
  archivePrefix = {arXiv},
  primaryClass  = {cs.LG}
}

@inproceedings{cao2026procache,
  title         = {{ProCache}: Constraint-Aware Feature Caching with Selective Computation
                   for Diffusion Transformer Acceleration},
  author        = {Cao, Fanpu and Chen, Yaofo and You, Zeng and Luo, Wei},
  booktitle     = {Proceedings of the AAAI Conference on Artificial Intelligence},
  volume        = {40},
  pages         = {19862--19870},
  year          = {2026},
  doi           = {10.1609/aaai.v40i24.39069},
  eprint        = {2512.17298},
  archivePrefix = {arXiv},
  primaryClass  = {cs.CV}
}

@inproceedings{zhang2025framepack,
  title={Frame Context Packing and Drift Prevention in Next-Frame-Prediction Video Diffusion Models},
  author={Zhang, Lvmin and Cai, Shengqu and Li, Muyang and Wetzstein, Gordon and Agrawala, Maneesh},
  booktitle={Advances in Neural Information Processing Systems},
  year={2025}
}

@misc{wang2026matrixgame3,
  title={{Matrix-Game 3.0}: Real-Time and Streaming Interactive World Model with Long-Horizon Memory},
  author={Wang, Zile and Liu, Zexiang and Li, Jiaxing and Huang, Kaichen and Xu, Baixin and Kang, Fei and An, Mengyin and Wang, Peiyu and Jiang, Biao and Wei, Yichen and Xietian, Yidan and Pei, Jiangbo and Hu, Liang and Jiang, Boyi and Xue, Hua and Wang, Zidong and Sun, Haofeng and Li, Wei and Ouyang, Wanli and He, Xianglong and Liu, Yang and Li, Yangguang and Zhou, Yahui},
  year={2026},
  eprint={2604.08995},
  archivePrefix={arXiv},
  url={https://arxiv.org/abs/2604.08995}
}

@inproceedings{wang2026precisecache,
  title={{PreciseCache}: Precise Feature Caching for Efficient and High-Fidelity Video Generation},
  author={Wang, Jiangshan and Zhao, Kang and Guo, Jiayi and Wang, Jiayu and Guo, Hang and Zhu, Chenyang and Li, Xiu and Yue, Xiangyu},
  booktitle={International Conference on Learning Representations},
  year={2026}
}

\clearpage
\appendix

\section{Extended Characterization}
\label{sec:appendix_characterization}
This appendix details the profiling protocol and the analyses summarized in
Sec.~\ref{sec:characterization}.

\subsection{Profiling Protocol}
\label{sec:appendix_char_protocol}
Unless noted, reuse error is the relative $L_2$ deviation
$\|\hat{\mathbf{y}}-\mathbf{y}\|_2/\|\mathbf{y}\|_2$ between the output produced with the
tested reuse policy ($\hat{\mathbf{y}}$) and the corresponding full-compute output
($\mathbf{y}$) at the same denoising step, prompt, and seed. SSIM is computed against the
same-seed vanilla video. Each study uses eight
profiling prompts (six for the step-dependence curve of
Fig.~\ref{fig:characterization}(a) and the age curves of
Fig.~\ref{fig:characterization_o4}) that are disjoint from the benchmark prompts.
All characterization runs use a reduced setting rather than the protocol of
App.~\ref{sec:appendix_impl}: $320\times544$ resolution, a single $81$-frame segment generated synchronously
(\texttt{ar\_step}$\,{=}\,0$: all frames share one noise level; no autoregressive rollout), $50$ steps, seed $0$, and the Diffusers releases of the checkpoints
(\texttt{SkyReels-V2-DF-\{14B,1.3B\}-540P-Diffusers}).

\subsection{Observation 1: Step Dependence and Schedule Comparison}
\label{sec:appendix_char_o1}
Along a full-compute trajectory, at every step $s$ and every transformer block we reuse
that block's residual from step $s-1$ given the block's exact input and measure the relative
error against the exact output, so errors never compound; the curve averages these
block-local errors over the $40$ blocks and $6$ prompts. ``Early'' and ``late'' in the $2.4\times$ ratio denote the mean error over the
steps $1$--$10$ and $40$--$49$ of the $50$-step trajectory, respectively.
The placement comparison (Fig.~\ref{fig:characterization}(b)) compares three placements of
exact refreshes: dense-to-sparse ($7$ refreshes, $2.62\times$; gaps increasing along the
trajectory), uniform ($8$ refreshes, $2.49\times$; equal gaps), and sparse-to-dense ($7$
refreshes, $2.62\times$; the mirror
image). Dense-to-sparse attains the highest mean SSIM, ahead of uniform on $6$ of the $8$
paired prompts (one tie) and of sparse-to-dense on $7$; the full-controller schedule
ablation on the development set is reported in
Sec.~\ref{sec:ablation_schedule}.

\subsection{Observation 2: Sparse Execution Details}
\label{sec:appendix_char_o2}
Two prototypes (Fig.~\ref{fig:characterization}(c)) are profiled at token fractions
$q\in\{0.10,0.15,0.20,0.30\}$ (Table~\ref{tab:appendix_sparse_exec}): a dense-forward token-refresh reference and true
sparse kernels.
\emph{Token refresh} is a dense-forward quality reference: at every step between anchors it runs the full
transformer forward for all tokens and keeps the fresh output only for the selected tokens (the others keep their
cached output). Its speedup is therefore flat at $1.23\times$ by construction (only the last ten of the $50$ steps are
free whole-step copies) and does not measure the cost of sparse execution; the speed argument rests on the true sparse
kernels.
\emph{True sparse kernels} execute attention and FFN only for the selected tokens; the
kernel is exact for those tokens, but non-selected tokens contribute keys and
values re-projected from their cached, stale hidden states. Its speedup rises to $1.94\times$ at $q=0.10$, still well below
its own compute-proportional projection ($2.72\times$), mainly because costs that do not
shrink with $q$ remain, chiefly re-projecting the keys and values of the full context (the
largest single item) and per-block launch overhead; the stale keys/values additionally incur an
$\approx0.024$ SSIM reduction relative to the dense-forward token-refresh reference run on the same six-anchor schedule (a separate run from the
three-anchor speed measurement of Table~\ref{tab:appendix_sparse_exec}). Whole-step output reuse with dense-to-sparse refreshes, which skips the transformer
entirely at reused steps, reaches $2.62\times$ under the same conditions. The deployed controller (Sec.~\ref{sec:framework_routing}) matches
neither prototype exactly: repaired tokens run as a gathered batch through every layer and attend to the cached keys and values of all
tokens without re-projecting them; its cost enters Sec.~\ref{sec:experiments} only through measured end-to-end latency.

\begin{table}[h]
\centering
\caption{\textbf{True sparse token repair stays far below its compute-proportional ideal, and whole-step dense-to-sparse
(D2S) reuse is faster.}
End-to-end speedup of the true sparse kernels and of the dense-forward token-refresh reference (flat by construction)
versus their compute-proportional ideals (for token refresh, the speedup its schedule would reach if only the selected
tokens were computed; SkyReels-V2 1.3B, profiling
setting of App.~\ref{sec:appendix_char_protocol}); the ideals differ because token refresh uses $3$ exact anchors and
true sparse kernels use $6$.}
\label{tab:appendix_sparse_exec}

\small
\setlength{\tabcolsep}{3pt}

\resizebox{\ifdim\width>\textwidth\textwidth\else\width\fi}{!}{%
\begin{tabular}{@{}lcccc@{}}
\toprule
Token budget $q$ & 0.10 & 0.15 & 0.20 & 0.30 \\
\midrule
Compute-proportional ideal (token refresh) & 3.18$\times$ & 2.93$\times$ & 2.72$\times$ & 2.37$\times$ \\
Compute-proportional ideal (true sparse) & 2.72$\times$ & 2.55$\times$ & 2.40$\times$ & 2.16$\times$ \\
Token refresh (dense-forward reference) & 1.23$\times$ & 1.23$\times$ & 1.23$\times$ & 1.23$\times$ \\
True sparse kernels & 1.94$\times$ & 1.83$\times$ & 1.78$\times$ & 1.65$\times$ \\
Whole-step D2S reuse (reference) & \multicolumn{4}{c}{2.62$\times$} \\
\bottomrule
\end{tabular}%
}
\end{table}

\subsection{Observation 3: Approximate-Anchor Variants}
\label{sec:appendix_char_o3}
An anchor step is the reference that all following non-anchor steps replay from; any
approximation inside it is inherited by every step until the next anchor. We tested
$17$ approximate-anchor policies grouped into three families:
(i) \emph{pyramid kernels}, which inside refresh steps reuse cached block residuals band by
band when per-block gates on input drift, residual stability, and age allow it, and reconstruct
skipped blocks from the cached bands (\emph{pyramid-full} is the complete kernel);
(ii) \emph{block-skip gates at anchors}, which skip transformer blocks whose input change
falls below a threshold even during refresh steps; and
(iii) adaptive anchor skipping, which skips a nominal anchor when the measured input risk
is low and keeps the whole cache instead.
Only three of these points lie above the clean-anchor Pareto frontier of
Fig.~\ref{fig:characterization}(d): block-skip gates at $3.01$--$3.04\times$ with identical
SSIM ($0.824$), up to $0.008$ above its linear interpolation and not dominated by any
clean schedule. The pyramid-full variant has approximately $0.046$ lower SSIM than the clean full-anchor baseline at the same
nominal budget. Block-skip gates at anchors provide additional speedup ($2.00$--$3.54\times$ vs.\ the clean full-anchor baseline's $1.94\times$) at lower SSIM
than the clean baseline.

\subsection{Observation 4: Frequency-Resolved Staleness}
\label{sec:appendix_char_o4}
\begin{figure}[t]
\centering
\includegraphics[width=\linewidth]{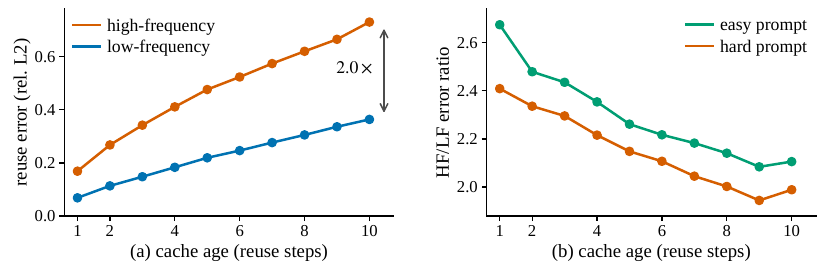}
\caption{
\textbf{High-frequency residual content goes stale faster than low-frequency content, but the separation is
content-dependent (Observation~4; SkyReels-V2 14B, six profiling prompts).}
\textbf{(a)} Relative-$L_2$ reuse error versus cache age (number of reuse steps since the cached state was refreshed)
for the low-frequency band of the block residuals (two levels of $2\times2$ average pooling) and the high-frequency
remainder, each measured against the corresponding full-compute band; HF error grows faster in absolute terms and
remains $\approx2$--$2.5\times$ the LF error across ages $1$--$10$ (up to $2.7\times$ for individual prompts; (b)).
\textbf{(b)} The HF/LF error ratio from (a) for the lowest-error and highest-error profiling prompts (``easy prompt'' and
``hard prompt'' in the legend); it stays above one throughout but is smaller on the highest-error prompt and shrinks with age.
}
\label{fig:characterization_o4}
\end{figure}

\paragraph{Measurement.}
On a full-compute trace, the exact residual $\mathbf{u}$ of every transformer block is
split on the token grid into a low-frequency band (two levels of $2\times2$ average
pooling, upsampled) and the high-frequency remainder. This probes where reuse error lives in
the residuals; the deployed controller instead measures latent drift with a single $2\times2$
pool (Eq.~\ref{eq:bands}). For each cache age $k\in\{1,\dots,10\}$ we replay the band of
the residual cached $k$ steps earlier and measure its relative-$L_2$ deviation from the exact
band
(Fig.~\ref{fig:characterization_o4}(a)). The profiling set contains six prompts;
the lowest-error and highest-error prompts, i.e., those with the lowest and highest mean reuse
error, are a static museum statue and a texture-rich koi pond, respectively. The HF/LF
ratio (Fig.~\ref{fig:characterization_o4}(b)) decreases from $2.67$ to $2.11$ on the lowest-error
prompt and from $2.41$ to $1.99$ on the highest-error prompt over ages $1{\to}10$. Pooling the
lower- and higher-error halves of the six prompts preserves this ordering at every age
($2.51{\to}2.04$ vs.\ $2.43{\to}1.98$), and HF stays above LF on all six prompts individually.

\paragraph{Rationale for Soft Frequency-Aware Control.}
Because the HF/LF separation is content-dependent, a design that assigns tokens to two
independently fixed anchor schedules according to their frequency content would be
mis-calibrated precisely on high-error prompts, where the gap between bands is smallest. FSC
therefore uses frequency content as a soft input to a per-step risk score
(Sec.~\ref{sec:design_risk}): high-frequency drift detects local detail and texture
change, low-frequency drift detects coarse structural shift, and cache age guards against
tokens with small per-step but large accumulated drift.

\section{FSC Design Details}
\label{sec:appendix_design}
This appendix expands Sec.~\ref{sec:design} with the full construction of the anchor
schedule, the complete definition of the staleness estimator, the exact routing rules
and cache-state updates, the backbone-specific instantiations, and the per-step
algorithm; Appendix~\ref{sec:appendix_oracle} validates the estimator against an oracle.

\paragraph{Notation.}
Throughout, transformer block $\ell$ at denoising step $s$ maps the hidden states
$\mathbf{H}^{\ell}_s$ of all $N$ tokens as
$\mathbf{H}^{\ell+1}_s=\mathbf{H}^{\ell}_s+\mathbf{U}^{\ell}_s$ with
$\mathbf{U}^{\ell}_s=U^{\ell}_\theta(\mathbf{H}^{\ell}_s,s,\mathbf{c})$, whose computation
dominates the cost; we write $\mathbf{h}^{\ell}_{s,i}=[\mathbf{H}^{\ell}_s]_i$ and
$\mathbf{u}^{\ell}_{s,i}=[\mathbf{U}^{\ell}_s]_i$ for the hidden state and the residual of token $i$.
Residual caching replays a stored $\mathbf{u}^{\ell}_{\mathrm{cache},i}$ instead, keeping the current hidden state.
We write residual replay blockwise for notation; the implementation stores the equivalent whole-stack residual per token,
while keys and values are cached per block.

\subsection{Anchor Schedule Construction}
\label{sec:appendix_design_schedule}
For $n$ post-warmup anchors there are $n-1$ inter-anchor gaps, to which FSC assigns
linearly increasing weights $\omega_i=1+i$, $i=0,\ldots,n-2$. The normalized cumulative
weights
\begin{equation}
c_0=0,\qquad
c_j=\frac{\sum_{i=0}^{j-1}\omega_i}{\sum_{i=0}^{n-2}\omega_i},\qquad j=1,\ldots,n-1,
\end{equation}
map the anchors onto $[W,E]$ via Eq.~\ref{eq:anchors}. A monotonic integer correction
then removes duplicate locations, enforces strict increase, and fixes the endpoints
$a_0=W$ and $a_{n-1}=E$. Degenerate cases ($n\le1$) reduce to at most one anchor; all
released configurations use $n\ge4$. A step is a clean refresh whenever
$\mathrm{anchor}(s)=(s<W)\vee(s\in\mathcal{A})$. Increasing gaps allocate more exact
computation to the early, less-cacheable part of the trajectory and exploit greater redundancy later.

\subsection{Staleness Estimator: Definitions}
\label{sec:appendix_design_risk}
\paragraph{Token alignment.}
Both bands of Eq.~\ref{eq:bands} are partitioned using the same local spatial layout as
the backbone's patch embedding, so each proxy token corresponds to exactly one
transformer token; $\mathbf{X}_s\in\mathbb{R}^{C\times F\times H_{\mathrm{lat}}\times W_{\mathrm{lat}}}$ denotes the
raw latent, and $\varepsilon$ in Eq.~\ref{eq:drift} is a small constant for numerical
stability.

\paragraph{Accumulated drift and age.}
The accumulated high-frequency drift is updated as
$A^{\mathrm{HF}}_{s,i}\leftarrow A^{\mathrm{HF}}_{s-1,i}+d^{\mathrm{HF}}_{s,i}$ and reset
to zero whenever token $i$ is recomputed; the age $a_{s,i}$ is incremented at every reuse
step and reset to zero on recompute. Because every $d^{\mathrm{HF}}_{s,i}$ is measured against the anchor reference
(Eq.~\ref{eq:drift}), $A^{\mathrm{HF}}$ is a time integral of anchor-relative displacement, not the path length
$\sum_s\|\mathbf{X}_{s}-\mathbf{X}_{s-1}\|$: a token displaced from the reference keeps accumulating even if it stops moving. Summing anchor-relative drift over the reuse duration, rather than step-to-step increments, thus penalizes sustained staleness.

\paragraph{Group-level aggregation.}
The instantaneous drifts are summarized by their $95$th percentiles,
$g^{\mathrm{B}}_s=Q_{0.95}(\{d^{\mathrm{B}}_{s,i}\}_{i=1}^{N})$, which emphasizes
localized stale regions while being less sensitive than the maximum to isolated noise;
the group age is $g^{\mathrm{age}}_s=\frac{1}{N}\sum_i a_{s,i}/k_{\mathrm{tok}}$. The
weights $w_{\mathrm{HF}}$, $w_{\mathrm{LF}}$, $w_{\mathrm{age}}$ in Eq.~\ref{eq:risk}
set the relative contribution of instantaneous detail drift, coarse structural drift,
and accumulated reuse age.

\paragraph{Properties of the proxy.}
First, $d^{\mathrm{B}}_{s,i}$ measures drift of the evolving latent relative to the
anchor latent; it is a proxy, not a direct measurement of residual error
(validated in Appendix~\ref{sec:appendix_oracle}). Second, for tokens repaired between
anchors, drift is still measured against the anchor reference, so their risk is
systematically over-estimated. This makes the routing decision conservative: over-estimation can only move the controller toward more
computation at the step level. Third, frequency is used as a soft
cache-validity signal rather than a permanent partition of tokens into independently
scheduled bands, because the HF/LF separation narrows modestly on higher-error prompts
(Fig.~\ref{fig:characterization_o4}(b)).

\subsection{Routing Rules and Repair Levels}
\label{sec:appendix_design_routing}
Let $\mathrm{viol}_s=\exists\,i:\;a_{s,i}\ge k_{\mathrm{tok}}$ indicate that at least one
token has reached its reuse-age limit, and let $\mathrm{skips}$ count consecutive
whole-step reuses. With the thresholds of Eq.~\ref{eq:thresholds}, a non-anchor step is
routed by evaluating the following cases top to bottom and applying the first match:
\begin{equation}
\mathrm{mode}(s)=
\begin{cases}
\mathrm{Full}, & \rho_s>t_{\mathrm{full}}
   \;\vee\;\big(\mathrm{viol}_s\wedge\rho_s>\eta\,t_{\mathrm{full}}\big),\\[3pt]
C0, & \rho_s<t_{\mathrm{skip}}\;\wedge\;\mathrm{skips}<k_{\mathrm{step}}
   \;\wedge\;\neg\mathrm{viol}_s,\\[3pt]
C1, & \rho_s<t_1,\\
C2, & \rho_s<t_2,\\
C3, & \text{otherwise.}
\end{cases}
\label{eq:routing}
\end{equation}
The relaxation factor $\eta<1$ lowers the $\mathrm{Full}$ bar whenever any token is overdue
(we use $\eta=0.8$ throughout), and $k_{\mathrm{step}}$ prevents indefinitely repeated
$C0$ decisions. The phase multipliers $\varphi_m(s)$ use tighter reuse thresholds early
in the trajectory and progressively looser thresholds later; exact base thresholds
$b_m$ and multipliers are listed in Appendix~\ref{sec:appendix_impl}.

The level-specific repair fractions are $q_m=q_{\mathrm{scale}}\,\hat q_m$ with
$\hat q_1<\hat q_2<\hat q_3$. At level $C3$ the optional low-frequency structural
safeguard augments Eq.~\ref{eq:repairset} as
$\mathcal{S}_s\leftarrow\mathcal{S}_s\cup
\mathrm{Top}_{p_{\mathrm{LF}}}(\{d^{\mathrm{LF}}_{s,i}\}_{i=1}^{N})$,
so that tokens with large coarse structural drift are repaired even when accumulated
high-frequency drift alone would not select them. The resulting block output is
\begin{equation}
\widehat{\mathbf{h}}^{\ell+1}_{s,i}=
\begin{cases}
\mathbf{h}^{\ell}_{s,i}+\mathbf{u}^{\ell}_{s,i}, & i\in\mathcal{S}_s,\\[3pt]
\mathbf{h}^{\ell}_{s,i}+\mathbf{u}^{\ell}_{\mathrm{cache},i}, & i\notin\mathcal{S}_s,
\end{cases}
\end{equation}
i.e., repaired tokens execute the full block while reused tokens bypass both attention
and non-attention residual computation.

\subsection{Cache-State Update}
\label{sec:appendix_design_state}
For each compatible controller context, FSC maintains the clean LF/HF references,
per-token ages $a_i$, accumulated drift $A^{\mathrm{HF}}_i$, cached transformer
residuals, and the consecutive-skip counter. The state evolves with the selected mode:
\begin{itemize}
\item \textbf{Scheduled clean anchor.} All tokens are recomputed exactly and their
residual caches rewritten; the LF/HF reference is replaced by the current exact latent;
$a_i\leftarrow0$, $A^{\mathrm{HF}}_i\leftarrow0$ for all $i$; $\mathrm{skips}\leftarrow0$.
\item \textbf{$C0$.} No residual is recomputed; all ages increase by one;
$\mathrm{skips}\leftarrow\mathrm{skips}+1$.
\item \textbf{$C1$--$C3$.} Tokens in $\mathcal{S}_s$ are gathered and recompute their
queries, keys, values, and feed-forward outputs in every block; their new keys/values
overwrite their slots in the block's key/value cache and their queries attend to the full
cached key/value set of the context window (entries of non-repaired tokens date from
their last recompute). Their residuals are overwritten with $a_i\leftarrow0$,
$A^{\mathrm{HF}}_i\leftarrow0$; reused tokens age by one; the anchor latent reference is
unchanged; $\mathrm{skips}\leftarrow0$.
\item \textbf{Risk-triggered full repair ($\mathrm{Full}$).} All residuals are recomputed and all
age/accumulation variables reset, but the anchor latent reference is unchanged;
$\mathrm{skips}\leftarrow0$.
\end{itemize}
In the implementation the anchor latent reference is rewritten by exactly one routine,
which is invoked only on scheduled anchor steps; the risk-triggered full repair shares the
full-recompute path (rewriting the residual cache state and resetting ages and
accumulated drift) but never invokes the reference-rewriting routine, so the reference is pinned to the last scheduled
anchor by construction.
Keeping one drift coordinate per anchor interval means the group risk does not reset
after a full repair, so it can re-trigger later in the same anchor
interval. On the
$946$-prompt benchmark FSC-Quality routes
$70.3\%$ / $12.0\%$ / $2.9\%$ / $8.9\%$ / $5.8\%$ of non-anchor steps to
$C0$ / $C1$ / $C2$ / $C3$ / $\mathrm{Full}$, and FSC-Speed
$69.1\%$ / $13.2\%$ / $2.8\%$ / $9.2\%$ / $5.7\%$.

\subsection{Backbone Instantiations}
\label{sec:appendix_design_backbones}
\paragraph{SkyReels-V2.}
SkyReels-V2 operates over spatial tokens and temporal groups within a diffusion-forcing
sliding window, and classifier-free guidance introduces separate conditional and
unconditional forward phases. FSC keeps independent controller state for each compatible
(phase, group) context, so cache age, accumulated drift, frequency references, and cached
residuals are never mixed across incompatible contexts, and group-local denoising steps determine both anchor
scheduling and phase-aware routing.

\paragraph{MAGI-1.}
MAGI-1 exposes chunk-level and token-level relative-$L_1$ reuse thresholds together with
a token-budget cap. FSC transfers to MAGI-1 through this native cache interface by scaling the token-level threshold,
\begin{equation}
\tau_{\mathrm{token}}=\frac{\beta\,r}{w_{\mathrm{HF}}}\qquad(w_{\mathrm{HF}}=1),
\end{equation}
where $\beta$ is the substrate's base reuse threshold and $r$ the threshold scale. The chunk-level threshold is read only
in MAGI's chunk-wise-only phase, which every FSC run disables, and the token-budget cap stays at its native value, so
the LF weight $w_{\mathrm{LF}}$ and the repair-fraction scale $q_{\mathrm{scale}}$ have no effect on MAGI-1.
The dense-to-sparse schedule runs on MAGI's global denoising-step
counter; at scheduled anchors the reuse thresholds are set to zero, forcing exact
computation, and the age-control principle is realized through a forced full recomputation
every $k_{\mathrm{tok}}$ steps of the same counter. Among the two primary Table~\ref{tab:main_results} backbones, the
full LF/HF-and-age controller runs on SkyReels-V2, while MAGI-1 tests transfer through its native cache
interface. The full controller is additionally evaluated end-to-end on
FramePack and Matrix-Game in App.~\ref{sec:appendix_crossfamily}.

\subsection{Complexity}
\label{sec:appendix_design_complexity}
Per denoising step, the SkyReels controller performs one $2\times2$ spatial average
pool, an upsampling and subtraction to obtain the two bands, and per-token norm and
ranking operations. The latent band operations are linear in the latent size, while percentile computation and
token ranking add an $\mathcal{O}(N\log N)$ per-group term in the token count $N$. No auxiliary network or additional
transformer pass is required. A GPU
micro-benchmark with the deployed shapes ($10{,}200$ tokens per group; $612$ routing decisions
per video, both guidance phases, weighted by the measured routing shares, plus one band
extraction per forward) gives about $0.23$~s per video, under $0.2\%$ of the $161$~s
generation latency; band extraction alone is below $0.05\%$. The measured acceleration
comes from skipping transformer residual computation through whole-step replay and selective
token reuse.

\subsection{Per-Step Algorithm}
\label{sec:appendix_design_algorithm}
Algorithm~\ref{alg:fsc} summarizes one FSC denoising step on SkyReels-V2, combining the
schedule of Sec.~\ref{sec:framework_schedule}, the estimator of
Sec.~\ref{sec:design_risk}, the routing of Eq.~\ref{eq:routing}, and the state
updates of Appendix~\ref{sec:appendix_design_state}.

\begin{algorithm}[t]
\caption{Frequency-Step Cache (FSC), one denoising step $s$}
\label{alg:fsc}
\begin{algorithmic}[1]
\Require latent $\mathbf{X}_s$; warmup $W$; anchor set $\mathcal{A}$; clean LF/HF
references; cached residuals; token ages $a$; accumulated drift $A^{\mathrm{HF}}$;
skip counter $\mathrm{skips}$
\If{$s<W$ \textbf{or} $s\in\mathcal{A}$} \Comment{clean anchor}
  \State recompute all tokens; store fresh residuals; rewrite LF/HF references
  \State $a\gets0$;\; $A^{\mathrm{HF}}\gets0$;\; $\mathrm{skips}\gets0$;\; \Return
\EndIf
\State compute LF/HF bands (Eq.~\ref{eq:bands}) and drifts
  $d^{\mathrm{LF}},d^{\mathrm{HF}}$ w.r.t.\ the anchor reference (Eq.~\ref{eq:drift})
\State $A^{\mathrm{HF}}\gets A^{\mathrm{HF}}+d^{\mathrm{HF}}$;\;
  compute $\rho_s$ (Eq.~\ref{eq:risk}) and thresholds $t_m(s)$ (Eq.~\ref{eq:thresholds})
\State route $\mathrm{mode}\in\{C0,C1,C2,C3,\mathrm{Full}\}$ (Eq.~\ref{eq:routing})
\If{$\mathrm{mode}=\mathrm{Full}$}
  \State recompute all tokens; refresh residuals;
    $a\gets0$;\; $A^{\mathrm{HF}}\gets0$;\; $\mathrm{skips}\gets0$
\ElsIf{$\mathrm{mode}=C0$}
  \State replay cached residuals for all tokens;\;
    $a\gets a+1$;\; $\mathrm{skips}\gets\mathrm{skips}+1$
\Else
  \State build $\mathcal{S}_s$ (Eq.~\ref{eq:repairset});
    add top-$p_{\mathrm{LF}}$ LF drifters if $\mathrm{mode}=C3$ and LF gate enabled
  \State recompute tokens in $\mathcal{S}_s$ and refresh their residuals; replay the rest
  \State reset $a,A^{\mathrm{HF}}$ for repaired tokens; $a\gets a+1$ for reused tokens;
    $\mathrm{skips}\gets0$
\EndIf
\end{algorithmic}
\end{algorithm}

\section{Oracle Diagnostic of the FSC Staleness Risk}
\label{sec:appendix_oracle}
We measure how strongly the deployed FSC staleness risk $\rho_s$
(Eq.~\ref{eq:risk}) is associated with the error incurred by residual reuse. This controlled
profiling experiment uses the SkyReels-V2 1.3B backbone with our development set (eight prompts;
App.~\ref{sec:appendix_impl}) and a
development configuration of the same family: the risk weights, $k_{\mathrm{tok}}$, $k_{\mathrm{step}}$,
thresholds, repair fractions, and phase shaping of App.~\ref{sec:appendix_impl}, but $W{=}7$,
$n{=}9$, $r{=}1.75$, $q_{\mathrm{scale}}{=}1.00$, and $p_{\mathrm{LF}}{=}0.1$. This yields
$4{,}896$ eligible non-anchor controller decisions. At each decision, we evaluate the exact residuals for oracle measurement. The probe leaves the anchor reference,
token ages, accumulated drift, and routing state unchanged, but overwrites the probed group's K/V cache. Simpler
predictors computed from the same decisions are informative too: the number of steps
since the last scheduled anchor alone reaches Spearman $0.865$ (ROC-AUC $0.913$), the age term alone $0.715$, and the HF
or LF drift term alone $0.859$ / $0.867$; $\rho_s$ adds $0.028$ Spearman over the elapsed-step predictor, and its
partial Spearman correlation with $E_s$ controlling for elapsed steps is $0.472$, against $0.174$ for elapsed steps
controlling for $\rho_s$. End-to-end quality and acceleration are evaluated separately
at benchmark scale in Sec.~\ref{sec:experiments}.

\begin{figure}[t]
\centering
\includegraphics{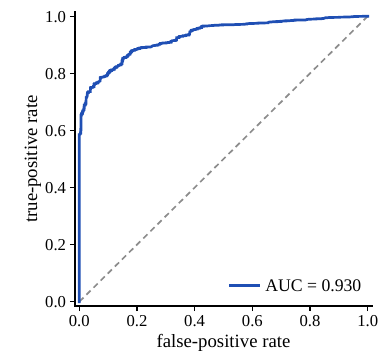}
\caption{\textbf{The FSC risk $\rho_s$ detects unsafe reuse ($E_s>0.05$) with ROC-AUC $0.930$.}
The $4{,}896$ non-anchor decisions of Fig.~\ref{fig:oracle_rho} (our development set, SkyReels-V2 1.3B, development
configuration described above). The oracle probe leaves the anchor reference and routing-state variables unchanged;
the K/V-cache side effect is described above. Dashed: chance.}
\label{fig:oracle_validation}
\end{figure}

For each eligible non-anchor decision $s$, let $\mathbf u^{\mathrm{exact}}_{s,i}$ denote
the exact current residual of token $i$, and let $\mathbf u^{\mathrm{cache}}_{s,i}$
denote the residual currently stored in FSC's cache. We define the token-level oracle
reuse error as
\begin{equation}
e_{s,i}
=
\frac{\left\|\mathbf u^{\mathrm{exact}}_{s,i}-\mathbf u^{\mathrm{cache}}_{s,i}\right\|_2}
     {\left\|\mathbf u^{\mathrm{exact}}_{s,i}\right\|_2+\varepsilon}.
\end{equation}
Because $\rho_s$ is a group-level routing signal, we aggregate token errors using the
same high-percentile principle as the FSC controller:
\begin{equation}
E_s = Q_{0.95}\!\left(\{e_{s,i}\}_{i=1}^{N}\right).
\end{equation}
The pooled Spearman correlation of Fig.~\ref{fig:oracle_rho} is lower than every per-prompt value because
prompts differ in their baseline error: at equal $\rho_s$, $E_s$ differs by up to $2\times$ across
prompts, and removing each prompt's median $\log E_s$ raises the pooled correlation to $0.933$.
The unsafe-reuse criterion $E_s>0.05$ (Sec.~\ref{sec:design_risk}) is fixed independently of the ROC analysis and is
not selected to maximize AUC; under it, $3{,}216$ of the $4{,}896$ decisions ($65.7\%$) are labeled unsafe (ROC curve
in Fig.~\ref{fig:oracle_validation}). The monotonic rise of the mean and median oracle error across deciles of $\rho_s$
(Fig.~\ref{fig:oracle_rho}(b)) is a calibration-style check that the association is not driven by a few extreme
decisions.

\section{Implementation Details}
\label{sec:appendix_impl}

\paragraph{Hardware and Software.}
All main-table experiments run on a single server with NVIDIA RTX~PRO
6000 Blackwell GPUs ($97{,}887$~MiB of reported device memory, corresponding to $102.6$~GB; GB denotes $10^9$ bytes
throughout; driver 570.211.01), PyTorch 2.11.0 with
CUDA 12.8, and bfloat16 inference.
Each video is generated on one GPU with batch size~1; timed runs are launched
only on GPUs with no other process, and runs later slowed by co-located jobs
are flagged post hoc (Timing Procedure below).
VBench-Long~\citep{huang2025vbenchpp} evaluation uses the official repository at commit
\texttt{45e79ec} in \texttt{long\_vbench\_standard} mode (each video
scored on the VBench-Long dimensions of its prompt set with the canonical short
prompt); the four GRiT-based semantic dimensions run in a separate
detectron2 environment.
The reported VBench aggregate uses a fixed subset of eleven dimensions: subject
consistency, motion smoothness and dynamic degree (subject-consistency prompts),
overall consistency, aesthetic quality and imaging quality (overall-consistency
prompts), scene and background consistency (scene prompts), and color, object
class and appearance style (their own prompt sets); every dimension is scored on
its complete VBench-Long prompt set ($505$ prompts in total) with the official
normalization and dimension weights, and, as in the official implementation, a
video in which the detector finds no object of the prompt is excluded from that
method's color mean.
On MAGI-1, speedups are per-prompt ratios over the vanilla-clean prompt pool after per-method latency auditing; VBench aggregates use the evaluated dimension subset on
$500$ prompts (five color prompts in which the detector finds the object for no method contribute nothing), and
fidelity uses the first ten prompts of each prompt set without removing Dev ($110$ prompts).
LPIPS uses the AlexNet variant~\citep{lpips}; SSIM and PSNR are computed
per frame against the same-seed vanilla video and averaged; for every method,
including the baselines, the reference is the decoded same-seed vanilla video.
Fidelity differences between methods use the same paired,
prompt-set-stratified bootstrap ($B{=}10{,}000$) as the VBench
differences: on the $102$-prompt subset, FSC-Quality$-$MotionCache is SSIM $+0.031$ $[+0.021,+0.042]$, PSNR
$+1.04$~dB $[+0.81,+1.26]$, and LPIPS $-0.024$ $[-0.034,-0.015]$, and FSC-Speed$-$MotionCache is SSIM $+0.019$
$[+0.008,+0.030]$, PSNR $+0.53$~dB $[+0.34,+0.73]$, and LPIPS $-0.005$ $[-0.015,+0.004]$. FSC-Quality outperforms
MotionCache on $72$ (SSIM), $83$ (PSNR), and $66$ (LPIPS) of the $102$ prompts, and FSC-Speed on $62$, $71$, and $44$.
On VBench (all $505$ evaluated Bench prompts) neither difference is resolved: FSC-Quality$-$MotionCache $+0.000$
$[-0.006,+0.006]$ and FSC-Speed$-$MotionCache $-0.001$ $[-0.006,+0.005]$.
Our development set (Secs.~\ref{sec:design_risk} and~\ref{sec:ablation}) consists of eight prompts, the first prompt of
each of eight benchmark prompt sets.

\paragraph{SkyReels-V2 Generation Settings.}
Checkpoint \texttt{Skywork/SkyReels-V2-DF-1.3B-540P}; resolution
$544\times960$ (540P); $177$ frames at $24$\,fps; $50$ denoising steps
(FlowUniPC scheduler, shift $8.0$); diffusion-forcing settings
\texttt{ar\_step}$=5$, \texttt{causal\_block\_size}$=5$,
\texttt{base\_num\_frames}$=97$, \texttt{overlap\_history}$=17$,
\texttt{addnoise\_condition}$=20$; classifier-free guidance $6.0$;
seed $1024$ for every method and prompt.

\paragraph{MAGI-1 Generation Settings.}
Checkpoint MAGI-1 4.5B (distilled); resolution $720\times720$; $168$ frames at $24$\,fps;
$64$ denoising steps; chunk-wise autoregression with the model's native
runtime; seed $1024$.
Each video runs as a single-GPU subprocess with vendor-recommended offload settings.

\paragraph{FLOPs Accounting.}
On SkyReels-V2 the executed-token trace of every transformer block is
recorded for all $946$ prompts after the token-cache warmup; the untraced
warmup and overlap-prefix forwards are full computations and are added for
every method, vanilla included (five full forwards per temporal group for FlowCache and
MotionCache, i.e., their warmup $4$ plus the cache-initializing step that follows it; two for FSC and the SeaCache port). PFLOPs follow from the
per-token cost of the block's linear layers and self-attention over the
group's own tokens, with the unconditional pass charged as the conditional
one; attention to cached context keys and cross-attention are not charged, so
SkyReels-V2 PFLOPs are comparable across methods but are not a complete count
and are not on the same basis as the MAGI-1 values.
The MAGI-1 runtime exposes no such trace, so we instrument its block stack
to count executed token-slots and attended query--key pairs per forward and
compare them against the vanilla schedule replayed for the same step.
The analytic vanilla cost is $140.6$~PFLOPs per video ($12{,}150$ tokens
per chunk, seven chunks, $64$ steps, a four-chunk denoising window with
the model's \texttt{noise2clean\_kvrange}; linear layers $55.9$, self-attention
$84.2$, cross-attention with caption projections $0.5$~PFLOPs), $1.2\%$ above the $139$~PFLOPs
reported by MotionCache for the identical setting, and the instrumented
counter reproduces it exactly ($1.000$) on a vanilla run.
Each cached method's PFLOPs is this vanilla cost times its measured executed
fraction: FlowCache on the full run, MotionCache and both FSC points on the
same instrumented $20$-prompt sample (both FSC points were instrumented
separately with the same counter; FSC-Quality's per-prompt latencies reproduce its
benchmark run within $1\%$).
Sparse forwards are charged for executed query tokens only, matching the
SkyReels-V2 convention; charging the full reassembled key/value length
instead changes the MAGI-1 fractions by less than $0.03$.

\paragraph{Timing Procedure.}
Latency is wall-clock time of the complete generation call, bracketed by
\texttt{torch.cuda.synchronize()}, measured per prompt.
Speedup is computed per prompt as the same-prompt vanilla latency
divided by the method latency on the same GPU model, then averaged over
prompts; vanilla latencies are measured once per prompt and cached.
Because latency is sensitive to co-located jobs on shared servers, all
timed runs are launched on otherwise-idle GPUs, and a post-hoc audit flags
any run whose latency leaves the per-configuration clean band
(the fastest-quintile latency $+2\%$; on idle GPUs the latency of a fixed workload varies by less than $1\%$). The
flagged SkyReels-V2 runs ($348$ of the $946$ runs of the $W{=}5$, $r{=}1.65$ operating point
of Table~\ref{tab:full_bench_all}, generated while other jobs shared its GPUs, and one FSC-Speed run)
were re-measured on idle GPUs, after which every vanilla, FlowCache, and FSC run lies inside its band and all speedups of
Table~\ref{tab:full_bench_all} average over all $946$ prompts. MotionCache's and SeaCache-AR's executed compute varies more
across prompts, which places $95$ and $530$ of their runs above this band; measured instead against the latency predicted
from each run's executed compute, no MotionCache run and five SeaCache-AR runs are more than $2\%$ slower, and excluding
those five would change SeaCache-AR's speedup by $0.001\times$. On MAGI-1, where cached latency depends strongly on the prompt, the
band is taken around each prompt's latency predicted from the other cached
methods; it flags $8$ of $4{,}730$ cached runs, none of them from the two FSC points of Table~\ref{tab:main_results}, and without it
FlowCache would print $1.485\times$.

\paragraph{Peak Memory.}
Table~\ref{tab:main_results} reports \texttt{torch.cuda.max\_memory\_allocated}
measured in a fresh process per method on an otherwise idle GPU: the pipeline
(DiT, text encoder, VAE) is loaded ($14.8$~GB), the peak counter is reset, and the generation call is
bracketed by \texttt{cuda.synchronize}; the value is identical on every
prompt tested (spread $<0.02$~GB). MAGI-1 values come from the benchmark runs
themselves (one subprocess per video, counter reset before model load).
On SkyReels-V2 1.3B the numbers are $53.7$ / $58.1$ / $58.9$ / $58.4$~GB
(vanilla / FlowCache / MotionCache / both FSC operating points), and $58.6$~GB for the chunk-wise SeaCache port of
App.~\ref{sec:appendix_matched} (recorded inside a benchmark worker process, so an upper bound).
The $4.4$--$5.2$~GB above vanilla is primarily shared token-caching state and auxiliary tensors inherited from the
common caching stack; FSC's additional controller state is small relative to this total.

\paragraph{Hardware Portability.}
Table~\ref{tab:hardware} repeats the SkyReels-V2 comparison on an H100 80GB under the same generation protocol and seed
(speedups against the H100's own vanilla latency and
fidelity against the H100's own same-seed vanilla output, since vanilla
videos differ across GPUs).
Vanilla is $1.22\times$ faster on the H100 ($300$ vs.\ $367$~s), but the
relative ordering is unchanged: at MotionCache-matched latency
($2.26$--$2.29\times$ on both cards) FSC has higher mean SSIM/PSNR and lower mean
LPIPS (higher SSIM on $7/10$ and $18/23$ prompts, lower LPIPS on $5/10$ and
$15/23$), and its peak memory differs from
MotionCache's by at most $0.52$~GB on both machines.
The H100 columns use two other configurations of the same controller: the $23$-prompt column uses FSC-Quality's anchor schedule
($W{=}6$, $n{=}10$) at MotionCache-matched latency with base thresholds $(0.12,0.20,0.30,0.70)$, $r{=}1.9$,
$q_{\mathrm{scale}}{=}0.8$ and $p_{\mathrm{LF}}{=}0.1$ (all other values as in FSC-Quality), and the $10$-prompt column uses
the $W{=}7$, $n{=}9$ configuration of Table~\ref{tab:full_bench_all}.

\begin{table}[h]
\centering
\caption{\textbf{At MotionCache-matched latency, FSC's fidelity advantage has the same sign on both GPUs.}
Same code on two GPUs (SkyReels-V2 1.3B, seed $1024$; speedup and
fidelity against the local vanilla; paired FSC$-$MotionCache differences
with win counts). RTX columns: the FSC-Quality point of Table~\ref{tab:main_results} ($946$ prompts; fidelity on its
$102$-prompt subset) and a separate $8$-prompt re-measurement: eight prompts of the $102$-prompt fidelity subset, one per
MotionCache-protocol dimension and disjoint from the development set; vanilla, released MotionCache ($\tau{=}0.10$) and FSC-Quality
were generated in one session on the same GPU, and fidelity is against that session's vanilla with the ablation-harness metrics
of Table~\ref{tab:ablation}, so this column is comparable neither with the $946$-prompt column nor with the development-set
Table~\ref{tab:mcsweep_dev}; RTX peak memory (fresh-process protocol above) was measured for the $946$-prompt column only; H100 columns: the $W{=}7$, $n{=}9$ configuration ($10$ prompts) and FSC-Quality's anchor schedule with
$q_{\mathrm{scale}}{=}0.8$ ($23$ prompts; see text); H100 peak memory comes from the benchmark worker processes, and in the
$23$-prompt column MotionCache ran after FSC in the same process and inherited FSC's per-window cache-clearing flag,
which lowers its resident memory ($58.4$ vs.\ $59.0$~GB) but not its outputs; $W$: warmup, $n$: anchors, $q_{\mathrm{scale}}$: repair-fraction
scale.}
\label{tab:hardware}

\small
\setlength{\tabcolsep}{3pt}

\resizebox{\ifdim\width>\textwidth\textwidth\else\width\fi}{!}{%
\begin{tabular}{@{}lcccc@{}}
\toprule
 & RTX PRO 6000 (946) & RTX PRO 6000 (8) & H100 (10) & H100 (23) \\
\midrule
Vanilla latency (s) & 367 & 367 & 300 & 300 \\
MotionCache speedup & 2.275$\times$ & 2.285$\times$ & 2.278$\times$ & 2.259$\times$ \\
FSC speedup & 2.275$\times$ & 2.274$\times$ & 2.280$\times$ & 2.277$\times$ \\
$\Delta$SSIM (FSC$-$MC) & $+0.031$ (72/102) & $+0.053$ (8/8) & $+0.021$ (7/10) & $+0.028$ (18/23) \\
$\Delta$LPIPS (FSC$-$MC) & $-0.024$ (66/102) & $-0.060$ (6/8) & $-0.009$ (5/10) & $-0.021$ (15/23) \\
Peak mem.\ vanilla / MC / FSC (GB) & 53.7 / 58.9 / 58.4 & -- & 53.7 / 59.0 / 58.5 & 53.7 / 58.4 / 58.5 \\
\bottomrule
\end{tabular}%
}
\end{table}
On MAGI-1 all three cached methods peak at $19.7$~GB against $31.8$~GB for
the vanilla runtime: the shared FlowCache/MotionCache runtime on which all
three are built drops nearly-clean chunks from the attention window
(\texttt{discard\_nearly\_clean\_chunk}, its released default), which
lowers the footprint for FlowCache, MotionCache, and FSC alike, and the
three cache states themselves are indistinguishable at this granularity.
FlowCache's KV-cache compression, one of its two components, is disabled in every
FlowCache row (the shared SkyReels-V2 runtime does not implement it; on MAGI-1 it is left
off, as for every other method), so these rows measure FlowCache's reuse policy alone.
Compression would lower FlowCache's memory below the shared baseline and may also reduce
its attention cost; Table~\ref{tab:main_results} credits neither effect.

\paragraph{FSC Hyperparameters (SkyReels-V2).}
\emph{FSC-Quality}: warmup $W=6$; $n=10$ dense-to-sparse anchors with last eligible step $E=39$, giving
$\mathcal{A}=\{6,7,8,10,13,17,21,27,32,39\}$ (Eq.~\ref{eq:anchors}); threshold scale $r=1.65$; base thresholds
$(b_{\mathrm{skip}},b_1,b_2,b_{\mathrm{full}})=(0.16,0.24,0.34,0.80)$
with phase shaping $\varphi$ that scales them by
$(0.7,0.7,0.8,0.9)$ for $s<20$, $1.0$ for $20\le s<30$, and
$(1.4,1.4,1.3,1.2)$ for $s\ge30$; risk weights
$(w_{\mathrm{HF}},w_{\mathrm{LF}},w_{\mathrm{age}})=(1.0,0.5,0.25)$;
maximum token age $k_{\mathrm{tok}}=8$; consecutive-replay cap
$k_{\mathrm{step}}=4$; level-specific repair fractions
$(\hat q_1,\hat q_2,\hat q_3)=(0.05,0.10,0.15)$ with $q_{\mathrm{scale}}=1.25$;
LF gate enabled with $p_{\mathrm{LF}}=0.2$; overdue relaxation $\eta=0.8$ (App.~\ref{sec:appendix_design_routing}).
\emph{FSC-Speed}: identical except $W=5$, hence $\mathcal{A}=\{5,6,7,10,13,16,21,26,32,39\}$, and $r=1.55$.
The LF band uses a $2\times2$ average pool (Eq.~\ref{eq:bands}); proxy
tokens use the backbone patch layout ($2{,}040$ tokens per latent frame;
a temporal group of five latent frames yields $10{,}200$ tokens per
decision); controller state is per (CFG phase, temporal group).

\paragraph{FSC Hyperparameters (MAGI-1).}
\emph{FSC-Quality}: warmup $4$; fifteen dense anchors
$\{6,8,10,12,14,17,20,24,28,32,37,42,48,54,60\}$ on MAGI's global denoising-step counter (a
backbone-specific discrete approximation of the dense-to-sparse schedule rather than
Eq.~\ref{eq:anchors} verbatim, which is why $a_0\neq W$ here); $r=1.25$ on the MAGI-native base threshold
$\beta=0.0065$ with $w_{\mathrm{HF}}=1.0$, and $k_{\mathrm{tok}}=4$.
\emph{FSC-Speed}: identical except $k_{\mathrm{tok}}=8$.
The twelve-anchor configuration of Table~\ref{tab:magi_points} uses anchors $\{7,8,10,12,16,20,26,32,39,46,54,62\}$
(gaps $1,2,2,4,4,6,6,7,7,8,8$), $r=1.0$ and $k_{\mathrm{tok}}=8$.
The configuration files also set $w_{\mathrm{LF}}$ ($0.5$; $0.8$ for the twelve-anchor configuration) and
$q_{\mathrm{scale}}$ ($1.0$; $1.25$ for FSC-Speed), which have no effect on MAGI-1
(App.~\ref{sec:appendix_design_backbones}).
These values map onto MAGI's native reuse thresholds as in App.~\ref{sec:appendix_design_backbones}.

\paragraph{FSC Hyperparameters (FramePack and Matrix-Game).}
FramePack: $W{=}2$, anchors $\{2,4,9,15,24\}$, $(k_{\mathrm{tok}},k_{\mathrm{step}}){=}(8,4)$,
$r{=}1.65$, $q_{\mathrm{scale}}{=}1.25$, phase-shaping boundaries at steps $10$ and $15$ of its $25$ steps (instead
of $20$ and $30$); Matrix-Game: $W{=}6$, anchors
$\{6,7,8,9,11,14,17,20,24,29,34,39\}$, $(k_{\mathrm{tok}},k_{\mathrm{step}}){=}(8,4)$,
$r{=}2.5$, $q_{\mathrm{scale}}{=}1.0$; all other values as in FSC-Quality on SkyReels-V2 (calibration in
App.~\ref{sec:appendix_crossfamily}).

\section{Baselines and Operating Points}
\label{sec:appendix_search}

\paragraph{Baselines and Their Reproduction.}
MotionCache is run from its official repository with the released
backbone-specific configuration (SkyReels-V2: threshold $0.1$, warmup $4$,
$K{=}6$, $\alpha{=}0.6$; MAGI-1: rel-$L_1$ $0.0065$, warmup $5$,
token-wise reuse on), on the same machine, seed, and prompts as all other
rows; the absolute speedups reported in the MotionCache paper correspond
to a different hardware and runtime setting and are therefore not
comparable to Table~\ref{tab:main_results}, whereas the ordering of
methods is preserved (verified on an H100 in Table~\ref{tab:hardware}).
FlowCache is reproduced inside the same shared FlowCache/MotionCache
codebase as its chunk-wise special case---the configuration the
MotionCache paper identifies as FlowCache ($K{=}17$ in its sweep, i.e., the coarse full
update applied throughout generation, without token-level selectivity): on SkyReels-V2 a group-level
reuse/recompute gate with FlowCache's threshold $0.1$ and MotionCache's warmup $4$; on
MAGI-1 the chunk-wise rel-$L_1$ policy with FlowCache's official VBench
setting (threshold $0.01$, warmup $5$) and token-wise reuse disabled.
FlowCache's KV-cache compression is disabled (App.~\ref{sec:appendix_impl}), so these rows
reproduce its reuse policy only.
The FlowCache paper reports larger speedups ($2.38\times$ on MAGI-1, $6.7\times$ on
SkyReels-V2) in its own evaluation setting and configuration. In our setting, the
official FlowCache repository's SkyReels-V2 stack without compression is $2.21\times$ faster
than its own vanilla on a probe prompt ($368$ vs.\ $166$\,s), close to our reproduction
($180$\,s, $2.04\times$) at comparable fidelity (SSIM $0.758$ vs.\ $0.756$); the gap to the
published SkyReels-V2 number therefore comes from differences in evaluation setting and configuration,
not from the reimplementation. We therefore evaluate all rows on a common stack because the official
stack's \emph{vanilla} output diverges from the MotionCache-fork vanilla at the same seed
(SSIM $0.92$ between the two references), which would make same-seed fidelity incomparable across rows.
ARCache~\citep{nan2026arcache}, which evaluates FramePack-F1, SkyReels-V2, and the original Matrix-Game, has no public
implementation as of September 2026 and reports results under different settings (e.g., 257 frames in 30 steps on
SkyReels-V2 and the GameWorld benchmark on Matrix-Game), so it is not included.

\begin{figure}[H]
\centering
\includegraphics[width=\textwidth]{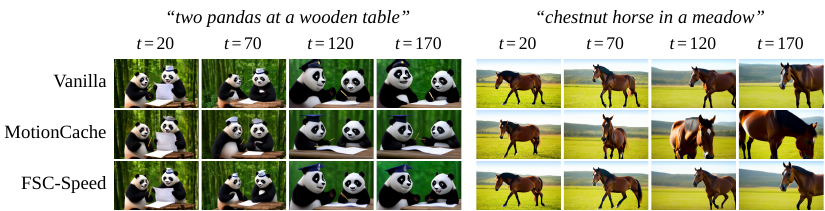}
\caption{
\textbf{FSC-Speed exhibits smaller deviation from the vanilla trajectory than MotionCache over time.}
SkyReels-V2 1.3B, two prompts, same seed; rows: vanilla, MotionCache, and FSC-Speed; columns: frames at four
timestamps. Early frames are nearly identical and deviations accumulate with time: MotionCache drifts in object
identity and layout (left: the cap morphs and the paper drifts away) and in framing and color (right: progressive
zoom and saturation).
}
\label{fig:qualitative}
\end{figure}

\subsection{Operating Points and Speed--Quality Frontier}
\label{sec:appendix_pareto}
\begin{figure}[t]
\centering
\includegraphics[width=\linewidth]{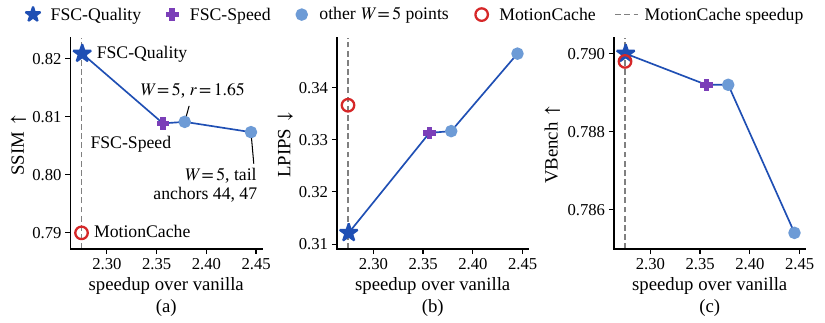}
\caption{
\textbf{Every FSC operating point has higher SSIM than MotionCache at equal or higher speed, with VBench within
$0.005$.}
SkyReels-V2, full benchmark (VBench on the fixed dimension subset of App.~\ref{sec:appendix_impl}; SSIM and LPIPS on
the $102$-prompt fidelity subset); FSC-Quality, FSC-Speed, and the two other $W{=}5$ points of
Table~\ref{tab:full_bench_all} (line: FSC points in speed order); the dashed line marks MotionCache's speedup.
}
\label{fig:pareto}
\end{figure}

Table~\ref{tab:full_bench_all} lists the SkyReels-V2 FSC operating points used elsewhere in the paper, and
Fig.~\ref{fig:pareto} plots FSC-Quality and the three faster $W{=}5$ points (FSC-Speed with $r{=}1.55$, $r{=}1.65$,
and $r{=}1.65$ with two tail anchors at steps $44$ and $47$) against MotionCache. Every FSC point improves PSNR and SSIM
over MotionCache at a VBench aggregate within $0.005$ of it while running $0$--$7.5\%$ faster, and all but the fastest
also improve LPIPS.
Dense-to-sparse anchors control the \emph{global} refresh frequency and
selective token repair recovers \emph{locally} stale content between
anchors, so FSC can spend computation continuously along the frontier
instead of choosing between full reuse and full recompute.

\begin{table}[t]
\centering
\caption{
\textbf{Every SkyReels-V2 FSC operating point improves PSNR and SSIM over MotionCache at a VBench aggregate within
$0.005$ of it.}
Full $946$-prompt benchmark, ordered by speedup (VBench on the fixed dimension subset of App.~\ref{sec:appendix_impl};
PSNR/SSIM/LPIPS on the $102$-prompt fidelity subset against the same-seed vanilla output). Configurations are
described relative to FSC-Quality ($W{=}6$, $n{=}10$ dense-to-sparse (D2S) anchors, $r{=}1.65$,
$q_{\mathrm{scale}}{=}1.25$, LF gate on with $p_{\mathrm{LF}}{=}0.2$). The two $W{=}5$ rows without tail anchors are
separate runs that differ only in $r$ ($1.55$ vs.\ $1.65$); the tail-anchor point adds exact steps $44$ and $47$ to the
$r{=}1.65$ row. Adding tail anchors changes the realized repair routing because anchors reset controller state:
the $W{=}5$, $r{=}1.65$ row routes $5.7\%$ of its non-anchor decisions to risk-triggered full repair, $2.9\%$ to $C2$ and
$8.6\%$ to $C3$; adding the tail anchors changes these shares to $0$, $0$ and $3.0\%$, hence the higher speed ($C0$: whole-step replay; $C1$--$C3$: token repair of increasing fraction; Full: full repair;
$W$: warmup; $n$: anchors; $r$: threshold scale). Best in \textbf{bold}, second best \underline{underlined}.
}
\label{tab:full_bench_all}

\small
\setlength{\tabcolsep}{3pt}

\resizebox{\textwidth}{!}{%
\begin{tabular}{@{}lccccc@{}}
\toprule
Configuration (relative to FSC-Quality) & Speedup$\uparrow$ & VBench$\uparrow$ &
PSNR$\uparrow$ & SSIM$\uparrow$ & LPIPS$\downarrow$ \\
\midrule
FSC-Quality (Table~\ref{tab:main_results}) & 2.275$\times$ & \textbf{0.7900} & \textbf{18.38} & \textbf{0.8209} & \textbf{0.3122} \\
$W{=}7$, $n{=}9$, $r{=}1.75$, $q_{\mathrm{scale}}{=}1.00$, $p_{\mathrm{LF}}{=}0.1$ & 2.277$\times$ & 0.7853 & \underline{18.31} & \underline{0.8202} & \underline{0.3199} \\
$W{=}5$, $r{=}1.55$ (FSC-Speed, Table~\ref{tab:main_results}) & 2.356$\times$ & 0.7892 & 17.87 & 0.8088 & 0.3313 \\
$W{=}5$ & \underline{2.378$\times$} & 0.7892 & 17.88 & 0.8091 & 0.3316 \\
$W{=}5$, tail anchors at steps $44$ and $47$ & \textbf{2.445$\times$} & 0.7854 & 17.79 & 0.8073 & 0.3464 \\
\midrule
MotionCache & 2.275$\times$ & \underline{0.7898} & 17.34 & 0.7899 & 0.3366 \\
\bottomrule
\end{tabular}%
}
\end{table}

\paragraph{MAGI-1 Operating Points.}
Table~\ref{tab:magi_points} lists the three FSC configurations evaluated on
the full MAGI-1 benchmark.
The three configurations trace different operating points of MAGI-1's native cache interface. The fifteen-anchor
configurations provide the strongest fidelity, while increasing $k_{\mathrm{tok}}$ from $4$ to $8$ gives a faster operating
point; Table~\ref{tab:main_results} reports the two fifteen-anchor points (FSC-Quality with $k_{\mathrm{tok}}{=}4$; FSC-Speed with
$k_{\mathrm{tok}}{=}8$). On the $110$ fidelity prompts, where every accelerated output has a same-seed vanilla reference, the VBench aggregate of every cached method (computed over the $60$ of these prompts that carry the evaluated dimensions) lies within the vanilla bootstrap CI (paired VBench differences: FSC-Quality$-$MotionCache $-0.009$ $[-0.030,+0.016]$, FSC-Quality$-$vanilla $+0.013$ $[-0.005,+0.032]$).
On this distilled backbone the opportunity for reuse between scheduled full recomputations is limited by the
$16$ denoising steps a chunk spends at each of the four window positions ($64$ steps per chunk in total), which is why the
FSC frontier on MAGI-1 spans a narrower speed range than on SkyReels-V2.

\begin{table}[h]
\centering
\caption{\textbf{On MAGI-1, every FSC configuration is faster than MotionCache with higher SSIM and lower LPIPS.}
Configurations evaluated over the full benchmark prompt pool; speedup uses the
vanilla-clean prompt pool after per-method latency auditing, VBench the $500$ prompts of the evaluated subset, fidelity the first ten
prompts of each prompt set ($110$; App.~\ref{sec:appendix_impl}). $\Delta$VBench is relative to MotionCache ($0.7950$); its paired bootstrap $95\%$ CI over prompts contains zero
for every row ($[-0.010,+0.004]$ for the twelve-anchor configuration, $[-0.007,+0.008]$ for
FSC-Quality, $[-0.013,+0.002]$ for FSC-Speed). $k_{\mathrm{tok}}$: forced full-recomputation period; $r$: threshold
scale (App.~\ref{sec:appendix_design_backbones}). Best in \textbf{bold}, second best
\underline{underlined}; the $\Delta$VBench marks rank the FSC rows only (MotionCache is the reference).}
\label{tab:magi_points}

\small
\setlength{\tabcolsep}{3pt}

\resizebox{\ifdim\width>\textwidth\textwidth\else\width\fi}{!}{%
\begin{tabular}{@{}lcccccc@{}}
\toprule
Configuration (relative to FSC-Quality) & Anchors & $k_{\mathrm{tok}}$ & Speedup$\uparrow$ & $\Delta$VBench$\uparrow$ &
SSIM$\uparrow$ & LPIPS$\downarrow$ \\
\midrule
$r{=}1.0$, 12 anchors, $k_{\mathrm{tok}}{=}8$ & 12 & 8 & 1.393$\times$ & \underline{$-0.0029$} & 0.8564 & 0.2722 \\
FSC-Quality (Table~\ref{tab:main_results}): dense anchors, $r{=}1.25$ & 15 & 4 & \underline{1.421$\times$} & $\mathbf{+0.0004}$ & \textbf{0.8883} & \textbf{0.2297} \\
FSC-Speed (Table~\ref{tab:main_results}): $k_{\mathrm{tok}}{=}8$ & 15 & 8 & \textbf{1.457$\times$} & $-0.0054$ & \underline{0.8782} & \underline{0.2460} \\
\midrule
MotionCache & -- & -- & 1.357$\times$ & -- & 0.8468 & 0.2778 \\
\bottomrule
\end{tabular}%
}
\end{table}

\subsection{Cross-Family Validation on FramePack and Matrix-Game}
\label{sec:appendix_crossfamily}
\emph{Backbones.} FramePack-F1 (\texttt{FramePack\_F1\_I2V\_HY\_20250503}, HunyuanVideo-based next-section
prediction; $25$ steps, distilled guidance $10$, $145$ frames at the $640$ bucket, seed $31337$, image-to-video from frame
$0$ of the SkyReels-V2 vanilla video) and the Matrix-Game-3.0 base model (Wan2.2-5B based; $50$ steps, CFG $5.0$, shift
$5.0$, $704\times1280$, $177$ frames, seed $42$; held-out prompts are likewise conditioned on frame $0$ of the SkyReels-V2
vanilla video).
\emph{Port.} The full controller runs with gathered-token execution as on SkyReels-V2 (selected tokens run every
block on a compact batch and attend to the full cached keys and values), with separate controller state per CFG branch.
\emph{Calibration.} FSC is calibrated per backbone on eight development prompts over $W$, $n$, $E$, $r$,
$q_{\mathrm{scale}}$ and $(k_{\mathrm{tok}},k_{\mathrm{step}})$; a configuration is eligible if its median latency is within
$5\%$ of the transferred SkyReels configuration, its LPIPS is not worse, and its PSNR is at most $0.1$~dB lower; the
eligible configuration with the highest SSIM is selected; the final configurations are listed in
App.~\ref{sec:appendix_impl}.
The coupled comparator is the purely reactive variant of Sec.~\ref{sec:ablation_separation}: the same
controller with warmup but no scheduled anchors, so risk-triggered full repairs also re-base the reference; its threshold scale is calibrated on the development prompts toward FSC's latency
($r{=}0.9$ and $0.543$).
\emph{Evaluation.} Each backbone is evaluated on $40$ held-out prompts stratified over the benchmark prompt sets and never
used for calibration; fidelity is SSIM/PSNR/LPIPS to the same-seed vanilla output (the implementation of
Table~\ref{tab:main_results}); CIs are stratified bootstrap ($B{=}10{,}000$) and wins are per-prompt.
\emph{Timing.} Latency ratios and speedups come from a same-GPU re-timing in which each prompt's controllers run
back to back on one idle GPU; vanilla, re-timed on five prompts per backbone, agrees with the original timing within
$0.2\%$. The re-timing reproduces the fidelity of the validation runs (mean $|\Delta\mathrm{SSIM}|\le0.006$ on
FramePack, $0$ on Matrix-Game), and its latency ratios agree with those of the validation runs, which used several GPUs,
to within $0.001$.
\emph{Transferred configuration.} Before calibration, the SkyReels-V2 FSC-Quality configuration was transferred by scaling its
step-count parameters to the sampler length, without tuning, and validated on the same $40$ prompts against a coupled
comparator (median latency ratios $1.02$ and $0.99$). FSC minus coupled was then
$\Delta$SSIM $+0.064$ $[+0.047, +0.083]$ and $\Delta$LPIPS $-0.057$ $[-0.069, -0.046]$ on Matrix-Game, and
$\Delta$SSIM $+0.026$ $[+0.018, +0.034]$ and $\Delta$PSNR $+1.46$~dB $[+0.82, +2.19]$ with an unresolved $\Delta$LPIPS
$-0.006$ $[-0.017, +0.004]$ on FramePack. Without calibration the separation result thus already holds on Matrix-Game in
all metrics and on FramePack in SSIM and PSNR but not LPIPS; calibration also improves FSC's own LPIPS (calibrated minus
transferred FSC: LPIPS $-0.022$ $[-0.031, -0.012]$ on FramePack and $-0.013$ $[-0.017, -0.009]$ on Matrix-Game).
Table~\ref{tab:crossfamily} reports the results. The coupled comparator is about $3\%$ faster than FSC on both backbones
(same-GPU median ratios $0.9685$ and $0.9667$), inside the pre-registered $[0.90, 1.10]$ band, and FSC is better
on both backbones in all three metrics. On
these two backbones, gathered token repair is close to compute-proportional (per-step cost divided by the
repaired fraction $0.69$--$0.87$ on FramePack and $0.58$--$0.96$ on Matrix-Game), unlike SkyReels-V2
(Observation~2).

\begin{table}[t]
\centering
\caption{\textbf{On FramePack and Matrix-Game, FSC outperforms a latency-matched coupled comparator in PSNR, SSIM, and
LPIPS, with every 95\% CI excluding zero.}
Cross-family validation on $40$ held-out prompts per backbone: FSC minus comparator, mean with stratified
bootstrap 95\% CI and prompt wins (FSC/comparator). Latency ratio is the same-GPU median per-prompt comparator/FSC latency
(each prompt's controllers run back to back on one idle GPU; bootstrap 95\% CI), and speedup is end-to-end relative to
the same-seed vanilla run; speedups are means of per-prompt ratios and can differ from the median latency ratio. Pre-registered matching criterion: median ratio in $[0.90, 1.10]$. Coupled: the purely reactive variant of Sec.~\ref{sec:ablation_separation} (the same controller without scheduled anchors, one risk signal triggering both repair and full synchronization); $r$: its threshold scale.}
\label{tab:crossfamily}
\resizebox{\linewidth}{!}{%
\begin{tabular}{lllcccc}
\toprule
Backbone & Comparator & Speedup FSC/comp.$\uparrow$ & Latency ratio & $\Delta$PSNR (dB) & $\Delta$SSIM & $\Delta$LPIPS \\
\midrule
FramePack & coupled ($r{=}0.9$) & 2.473$\times$/2.513$\times$ & \begin{tabular}[c]{@{}c@{}}0.9685\\[-1pt]{\scriptsize $[0.9676, 0.9707]$}\end{tabular} & \begin{tabular}[c]{@{}c@{}}$+5.18$\\[-1pt]{\scriptsize $[+4.03, +6.35]$, 40/0}\end{tabular} & \begin{tabular}[c]{@{}c@{}}$+0.079$\\[-1pt]{\scriptsize $[+0.065, +0.094]$, 40/0}\end{tabular} & \begin{tabular}[c]{@{}c@{}}$-0.092$\\[-1pt]{\scriptsize $[-0.105, -0.080]$, 39/1}\end{tabular} \\[3pt]
\midrule
Matrix-Game & coupled ($r{=}0.543$) & 2.445$\times$/2.535$\times$ & \begin{tabular}[c]{@{}c@{}}0.9667\\[-1pt]{\scriptsize $[0.9629, 0.9692]$}\end{tabular} & \begin{tabular}[c]{@{}c@{}}$+2.77$\\[-1pt]{\scriptsize $[+2.53, +3.00]$, 40/0}\end{tabular} & \begin{tabular}[c]{@{}c@{}}$+0.058$\\[-1pt]{\scriptsize $[+0.043, +0.075]$, 36/4}\end{tabular} & \begin{tabular}[c]{@{}c@{}}$-0.057$\\[-1pt]{\scriptsize $[-0.067, -0.047]$, 39/1}\end{tabular} \\[3pt]
\bottomrule
\end{tabular}}
\end{table}

\subsection{Matched-Latency Comparison with FlowCache}
\label{sec:appendix_matched}
FlowCache's released operating point ($2.04\times$) is $12\%$ slower than
FSC-Quality, whose latency equals MotionCache's.
To compare at equal latency in both directions we (i) sped FlowCache up to
FSC-Quality's latency by raising its reuse threshold ($0.10\!\to\!0.123$, chosen by
a three-prompt latency calibration) and (ii) slowed FSC down to FlowCache's
latency by adding anchors to FSC-Quality and lowering its threshold scale ($13$ D2S
anchors, $r{=}1.40$; all other hyperparameters unchanged, same calibration
procedure), and evaluated everything on the $235$-prompt stratified subset
on which all methods have results, giving a paired VBench comparison on the subset's $126$
prompts that carry the evaluated dimensions ($124$ for the slower block; Table~\ref{tab:matched_fc}).
At FSC-Quality's latency FlowCache's VBench aggregate ($0.7922$) stays within the CI of
FSC-Quality's ($+0.004$ $[-0.003,+0.011]$) and of MotionCache's ($+0.004$ $[-0.004,+0.012]$),
while its fidelity is far below FSC-Quality's (SSIM $0.772$ vs.\ $0.828$).
The slower FSC point has the highest SSIM and PSNR ($0.860$, $20.07$~dB; SeaCache's LPIPS is
lower), yet its VBench aggregate is $0.012$ below
FlowCache's (bootstrap $95\%$ CI $[-0.022,-0.003]$) and $0.004$ below
FSC-Quality on the $124$ prompts both were scored on (Table~\ref{tab:mcsweep_heldout}).
Additional repair computation therefore improves same-seed fidelity without a corresponding VBench change: the
fast operating points lie within one VBench bootstrap CI of each other, whereas the fidelity ordering is unambiguous.
The same table includes SeaCache~\citep{chung2026seacache}, which has no
autoregressive implementation; we port its scheduling rule onto the same
stack in the chunk-wise form of FlowCache (a group-level reuse/recompute gate driven by the
accumulated rel-$L_1$ of the SEA-filtered block-0
input, filter coefficients from each group's own noise level, first two steps and
the last step exact, its threshold $\delta$ calibrated to the two latencies).
At FSC-Quality's latency SeaCache behaves like MotionCache (SSIM $0.783$, VBench $0.791$ on
this subset; in Table~\ref{tab:main_results} the same $\delta{=}0.28$ point
gives $2.320\times$, VBench $0.7886$, SSIM $0.7919$, LPIPS $0.3331$); at the slower latency it matches the slow FSC point on fidelity (SSIM $0.854$ vs.\ $0.860$, but lower LPIPS, $0.236$ vs.\ $0.244$, on the
$21$ prompts with references) with a higher aggregate
($+0.006$ $[+0.0001,+0.012]$).
FSC's fidelity advantage over the spectral SeaCache-AR scheduler
is thus largest at the faster operating point, which is the regime the
matched-speed comparison of Table~\ref{tab:main_results} targets; over FlowCache it is larger at the slower point.

\begin{table}[h]
\centering
\caption{\textbf{At both latencies, the FSC point has the highest PSNR and SSIM.}
Paired comparison on the $235$-prompt stratified subset
(SkyReels-V2, seed $1024$; VBench on the $126$ subset prompts carrying the evaluated
dimensions, $124$ for the slower block, with bootstrap $95\%$ CIs over prompts; fidelity on the
$21$ prompts of the subset that have vanilla references). SeaCache-AR is
our chunk-wise port of SeaCache's scheduling rule (see text); thr, $\delta$: FlowCache and SeaCache-AR reuse
thresholds; $r$: FSC threshold scale. Best in \textbf{bold}, second best \underline{underlined} within each block.}
\label{tab:matched_fc}

\small
\setlength{\tabcolsep}{3pt}

\resizebox{\ifdim\width>\textwidth\textwidth\else\width\fi}{!}{%
\begin{tabular}{@{}lcccccc@{}}
\toprule
Method & Latency (s)$\downarrow$ & Speedup$\uparrow$ & VBench$\uparrow$ & PSNR$\uparrow$ & SSIM$\uparrow$ & LPIPS$\downarrow$ \\
\midrule
\multicolumn{7}{l}{\emph{faster operating point}}\\
MotionCache                  & 161 & 2.274$\times$ & 0.7885 [0.758, 0.817] & 17.55 & 0.7789 & 0.3156 \\
FlowCache (thr $0.123$)      & \underline{159} & \underline{2.305$\times$} & \textbf{0.7922} [0.762, 0.821] & 17.46 & 0.7715 & \underline{0.3125} \\
SeaCache-AR ($\delta{=}0.28$) & \textbf{158} & \textbf{2.321$\times$} & \underline{0.7908} [0.761, 0.820] & \underline{17.68} & \underline{0.7826} & 0.3222 \\
FSC-Quality                  & 161 & 2.275$\times$ & 0.7883 [0.757, 0.817] & \textbf{18.87} & \textbf{0.8277} & \textbf{0.2840} \\
\midrule
\multicolumn{7}{l}{\emph{slower operating point}}\\
FlowCache (thr $0.10$, released) & \textbf{180} & \textbf{2.037$\times$} & \textbf{0.7916} [0.759, 0.821] & 17.80 & 0.7825 & 0.2891 \\
SeaCache-AR ($\delta{=}0.22$) & 188 & 1.947$\times$ & \underline{0.7855} [0.752, 0.816] & \underline{19.82} & \underline{0.8539} & \textbf{0.2357} \\
FSC-Quality, $13$ anchors, $r{=}1.40$ & \underline{182} & \underline{2.015$\times$} & 0.7796 [0.746, 0.810] & \textbf{20.07} & \textbf{0.8600} & \underline{0.2435} \\
\bottomrule
\end{tabular}%
}
\end{table}

\subsection{Seed Robustness}
\label{sec:appendix_seeds}
We repeated the SkyReels-V2 comparison with four additional seeds ($7$,
$42$, $123$, $2026$) on a $78$-prompt subset stratified over the benchmark
prompt sets, generating vanilla, MotionCache, and FSC-Quality for every seed and
comparing each accelerated video with the vanilla video of the \emph{same}
seed (Table~\ref{tab:seeds}).
Fidelity margins replicate at every seed and are significant on the $78$
paired prompts of each added seed (SSIM $+0.021$ to $+0.044$, PSNR $+0.9$ to $+1.7$~dB, LPIPS
$-0.016$ to $-0.043$; paired $t$-test $p\le1.6\times10^{-5}$ for SSIM and PSNR and
$p\le0.022$ for LPIPS, the largest values both at seed $42$; FSC wins on $52$--$71$ of $78$ prompts; on the
$13$ prompts of seed $1024$, $p=0.044$ / $0.001$ / $0.098$ for SSIM / PSNR / LPIPS); over
the five seeds the mean margin is SSIM $+0.035\pm0.009$, PSNR
$+1.25\pm0.31$~dB, LPIPS $-0.031\pm0.011$, at identical speed
($2.274\pm0.002\times$ for MotionCache, $2.274\pm0.001\times$ for FSC;
per-seed ratio $1.000\pm0.001$).
The seed-to-seed standard deviation of the VBench aggregate is
$0.003$--$0.008$ per method; on this
subset FSC is $0.008\pm0.009$ below MotionCache on average, from $-0.019$
(seed $1024$) to $+0.003$ (seed $42$), whereas on the full benchmark the
same difference is $+0.000$ (paired bootstrap $95\%$ CI $[-0.006,+0.006]$)---the
$78$-prompt subset weights the evaluated dimensions differently from the full
benchmark, and MotionCache
scores above vanilla on it at two of the five seeds.
These results indicate that the fidelity advantage persists across seeds, whereas the VBench ranking between
accelerated methods remains sensitive to prompt-subset composition---differences of ${\sim}0.01$
change sign and size with the subset.
An additional $98$-prompt pool (subject-consistency, overall-consistency, and scene prompts) gives the same
fidelity ordering: FSC$-$MotionCache SSIM $+0.033$ at seed $7$ and $+0.034$ at seed $42$.

\begin{table}[h]
\centering
\caption{\textbf{FSC-Quality's fidelity advantage over MotionCache replicates at all five seeds at identical speed.}
Seed robustness on the $78$-prompt subset stratified over the
benchmark prompt sets (SkyReels-V2). VBench aggregate per seed for all three
methods; speedup of MotionCache / FSC-Quality over the same-seed vanilla; same-seed
fidelity difference FSC-Quality$-$MotionCache with the number of prompts on which
FSC-Quality wins; $n$ = prompts with a same-seed vanilla reference ($78$ per added seed; for seed $1024$ the $13$-prompt
overlap with the fidelity subset, scored with the in-pipeline metrics rather than the offline metrics of the added seeds). Mean $\pm$ sd rows use the unrounded per-seed values.}
\label{tab:seeds}

\small
\setlength{\tabcolsep}{3pt}

\resizebox{\ifdim\width>\textwidth\textwidth\else\width\fi}{!}{%
\begin{tabular}{@{}l cc ccc c ccc@{}}
\toprule
 & \multicolumn{2}{c}{Speedup} & \multicolumn{3}{c}{VBench ($41$ of the $78$ prompts)} & & \multicolumn{3}{c}{FSC $-$ MotionCache, same-seed fidelity} \\
Seed & MC & FSC & vanilla & MC & FSC & $n$ & $\Delta$PSNR (wins) & $\Delta$SSIM (wins) & $\Delta$LPIPS (wins) \\
\midrule
1024 & 2.274 & 2.274 & 0.7802 & 0.7892 & 0.7698 & 13 & $+1.06$ (11/13) & $+0.040$ (9/13) & $-0.028$ (9/13) \\
7 & 2.275 & 2.275 & 0.7748 & 0.7829 & 0.7736 & 78 & $+1.71$ (71/78) & $+0.044$ (63/78) & $-0.043$ (57/78) \\
42 & 2.275 & 2.274 & 0.7875 & 0.7851 & 0.7879 & 78 & $+0.91$ (55/78) & $+0.021$ (55/78) & $-0.016$ (52/78) \\
123 & 2.271 & 2.273 & 0.7898 & 0.7826 & 0.7835 & 78 & $+1.38$ (65/78) & $+0.036$ (65/78) & $-0.039$ (56/78) \\
2026 & 2.276 & 2.275 & 0.7855 & 0.7827 & 0.7701 & 78 & $+1.20$ (69/78) & $+0.033$ (67/78) & $-0.028$ (55/78) \\
\midrule
mean $\pm$ sd & $2.274\pm0.002$ & $2.274\pm0.001$ & $0.7836\pm0.0060$ & $0.7845\pm0.0028$ & $0.7770\pm0.0082$ &  & $+1.25\pm0.31$ & $+0.035\pm0.009$ & $-0.031\pm0.011$ \\
\bottomrule
\end{tabular}%
}
\end{table}

\subsection{MotionCache Threshold Sweep}
\label{sec:appendix_mcsweep}
\begin{figure}[t]
\centering
\includegraphics[width=\linewidth]{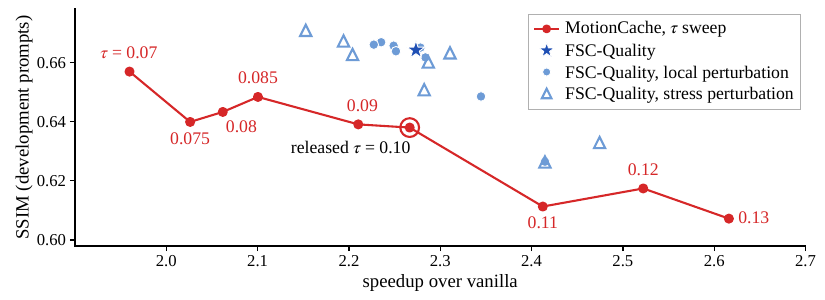}
\caption{\textbf{On the development prompts, every perturbed FSC-Quality point has higher SSIM than the MotionCache
threshold sweep at equal latency.}
SkyReels-V2, our development set (eight prompts); MotionCache's $\tau$ sweep of Table~\ref{tab:mcsweep_dev} (released
$\tau{=}0.10$ marked), FSC-Quality, and its local and stress perturbations of App.~\ref{sec:appendix_sensitivity}.}
\label{fig:mc_sweep_dev}
\end{figure}

Table~\ref{tab:main_results} runs MotionCache with its released configuration. To test whether this favors FSC-Quality, we sweep
MotionCache over its reuse threshold $\tau$, the parameter its release documents as controlling reuse aggressiveness (a
token is recomputed once its accumulated motion-weighted distance reaches $\tau$); warmup $4$, $K{=}6$, $\alpha{=}0.6$ and
all other released settings are kept. \emph{Development sweep.} On our development set (same stack, seed, GPU and metrics), nine thresholds span $1.96$--$2.62\times$
(Table~\ref{tab:mcsweep_dev}, Fig.~\ref{fig:mc_sweep_dev}); the released $\tau{=}0.10$ lies at FSC-Quality's latency, and at equal
latency every FSC point of App.~\ref{sec:appendix_sensitivity} has higher SSIM than the MotionCache sweep ($+0.015$ to
$+0.034$). \emph{Operating-Point Selection.} Before the evaluation below, three operating points were frozen from the development latencies alone:
the threshold closest to FlowCache's speed ($2.04\times$), the one closest to FSC-Quality's latency (within $3\%$), and the one
$5$--$8\%$ faster than FSC-Quality, giving $\tau{=}0.075$, $0.10$ and $0.11$. The latency-matched point is the released
threshold. \emph{Evaluation.} The frozen points are evaluated on the $235$-prompt stratified subset of
App.~\ref{sec:appendix_matched} (all outside the threshold-sweep set; VBench on $124$: the subset's $126$ prompts carrying the evaluated dimensions minus two scene prompts
without a $13$-anchor score; fidelity on the $21$ prompts that have a vanilla reference video and FSC-Quality fidelity), with the
paired bootstrap of App.~\ref{sec:appendix_matched}. The regenerated $\tau{=}0.10$ point matches the benchmark's
MotionCache row to within $0.003$ on VBench ($+0.0029$ $[+0.0001, +0.0063]$, a resolved but small difference). At matched latency FSC-Quality keeps its fidelity advantage (SSIM
$+0.049$ on $16/21$ prompts, PSNR $+1.3$~dB, LPIPS $-0.032$) while the VBench difference is unresolved
(MotionCache $-$ FSC-Quality $+0.0030$ $[-0.0070, +0.0136]$); the fidelity advantage also holds against the faster and the slower
frozen points (Table~\ref{tab:mcsweep_heldout}). On VBench, the faster MotionCache point scores
below FSC-Quality ($-0.013$ $[-0.023, -0.004]$), and the slower one has a higher aggregate than the slowed FSC point ($+0.013$ $[+0.004, +0.024]$). The
mechanically selected latency-matched point is the released threshold itself, so Table~\ref{tab:main_results} compares
MotionCache at its own latency-matched operating point.
Fig.~\ref{fig:mc_frontier}(a) plots every method on these $21$ fidelity prompts, with speedups averaged over the $235$ prompts.

\begin{table}[t]
\centering
\caption{\textbf{On the development prompts, MotionCache's released $\tau{=}0.10$ lies at FSC-Quality's latency, where
FSC-Quality has higher SSIM.}
MotionCache threshold sweep on our development set (eight prompts; seed $1024$, one idle GPU). Paths: share of
group decisions that reuse / fully update / selectively update. $^{s,m,f}$: frozen
slow, matched and fast points. The FSC-Quality row is run~1 of App.~\ref{sec:appendix_sensitivity};
Table~\ref{tab:ablation} lists the stored reference run (LPIPS $0.2925$). s/video: median wall-clock. Best in
\textbf{bold}, second best \underline{underlined}.}
\label{tab:mcsweep_dev}
\resizebox{0.92\linewidth}{!}{%
\begin{tabular}{lrrrrrc}
\toprule
$\tau$ & s/video$\downarrow$ & Speedup$\uparrow$ & PSNR$\uparrow$ & SSIM$\uparrow$ & LPIPS$\downarrow$ & Paths (\%) \\
\midrule
0.07 & 188 & 1.960$\times$ & \underline{17.89} & \underline{0.6569} & \textbf{0.2806} & 54/33/13 \\
0.075$^{s}$ & 181 & 2.026$\times$ & 17.67 & 0.6399 & 0.3040 & 54/28/18 \\
0.08 & 177 & 2.062$\times$ & 17.71 & 0.6433 & 0.3073 & 54/26/20 \\
0.085 & 175 & 2.100$\times$ & \underline{17.89} & 0.6483 & 0.2960 & 54/27/19 \\
0.09 & 165 & 2.210$\times$ & 17.49 & 0.6391 & 0.3058 & 62/27/11 \\
0.1$^{m}$ & 162 & 2.267$\times$ & 17.42 & 0.6380 & 0.3113 & 61/23/16 \\
0.11$^{f}$ & 152 & 2.412$\times$ & 17.17 & 0.6113 & 0.3540 & 64/22/14 \\
0.12 & \underline{146} & \underline{2.522$\times$} & 17.46 & 0.6174 & 0.3447 & 67/22/11 \\
0.13 & \textbf{140} & \textbf{2.616$\times$} & 16.47 & 0.6072 & 0.3564 & 70/21/9 \\
\midrule
FSC-Quality & 161 & 2.273$\times$ & \textbf{18.92} & \textbf{0.6641} & \underline{0.2926} & -- \\
\bottomrule
\end{tabular}}
\end{table}

\begin{table}[t]
\centering
\caption{\textbf{FSC keeps its fidelity advantage over every frozen MotionCache point outside the sweep set.}
Frozen MotionCache operating points on the $235$ prompts outside the threshold-sweep set (VBench on the $124$ carrying
the evaluated dimensions, with bootstrap 95\% CI; PSNR/SSIM/LPIPS on the $21$ paired fidelity prompts). FSC rows are the
stored results of App.~\ref{sec:appendix_matched}. Parentheses: fidelity margin of the nearest-latency FSC point (the
$13$-anchor point for $\tau{=}0.075$, FSC-Quality otherwise), computed from unrounded means, and FSC wins / MotionCache wins on the $21$ prompts; paired
VBench differences: $+0.0134$ $[+0.0037, +0.0240]$, $+0.0030$ $[-0.0070, +0.0136]$, $-0.0127$ $[-0.0227, -0.0035]$
(MotionCache $-$ FSC). s/video: median wall-clock; $\tau$: MotionCache reuse threshold; $r$: threshold scale. Best in \textbf{bold}, second best \underline{underlined}.}
\label{tab:mcsweep_heldout}
\resizebox{0.92\linewidth}{!}{%
\begin{tabular}{lrrcrrr}
\toprule
Method & s/video$\downarrow$ & Speedup$\uparrow$ & VBench$\uparrow$ & PSNR$\uparrow$ & SSIM$\uparrow$ & LPIPS$\downarrow$ \\
\midrule
MotionCache $\tau{=}0.075$ (slow) & 180 & 2.035$\times$ & \textbf{0.7930} {\scriptsize$[0.761, 0.823]$} & 17.84 {\scriptsize(FSC $+2.22$)} & 0.7869 {\scriptsize(FSC $+0.073$, 17/4)} & 0.2911 {\scriptsize(FSC $-0.048$, 13/8)} \\
MotionCache $\tau{=}0.10$ (matched; released value) & \underline{161} & \underline{2.278$\times$} & \underline{0.7870} {\scriptsize$[0.756, 0.816]$} & 17.55 {\scriptsize(FSC $+1.32$)} & 0.7784 {\scriptsize(FSC $+0.049$, 16/5)} & 0.3158 {\scriptsize(FSC $-0.032$, 15/6)} \\
MotionCache $\tau{=}0.11$ (fast) & \textbf{151} & \textbf{2.425$\times$} & 0.7713 {\scriptsize$[0.740, 0.801]$} & 17.29 {\scriptsize(FSC $+1.58$)} & 0.7718 {\scriptsize(FSC $+0.056$, 19/2)} & 0.3389 {\scriptsize(FSC $-0.055$, 20/1)} \\
\midrule
FSC-Quality & \underline{161} & 2.275$\times$ & 0.7840 {\scriptsize$[0.752, 0.814]$} & \underline{18.87} & \underline{0.8277} & \underline{0.2840} \\
FSC-Quality, $13$ anchors, $r{=}1.40$ & 182 & 2.015$\times$ & 0.7796 {\scriptsize$[0.746, 0.810]$} & \textbf{20.07} & \textbf{0.8600} & \textbf{0.2435} \\
\bottomrule
\end{tabular}}
\end{table}

\subsection{Comparison with SCOPE on Its Official Stack}
\label{sec:appendix_scope}
SCOPE~\citep{scope2026} is concurrent work whose official implementation (commit \texttt{05dc4dd}) provides an adapter for
SkyReels-V2-DF-1.3B on the upstream SkyReels-V2 runtime (revision \texttt{9351d13}, verified with the authors' checker), not
on the shared runtime of Table~\ref{tab:main_results}. We run it unmodified with the authors' released SkyReels-V2
configuration ($\tau_c{=}0.18$, $\tau_p{=}0.42$, second-order prediction, decay $0.75$, skip cap $5$, selective
computation and sensitivity guard on), loaded from the configuration file written by their own command line, and with our
benchmark protocol for everything else: the same checkpoint, $946$ prompts, seed, resolution, $177$ frames, window, overlap,
$50$ steps, guidance, shift and negative prompt, the same GPU model on idle devices, the same video writer, VBench-Long
evaluation and fidelity implementation. No SCOPE parameter was tuned.

\emph{Cross-Stack Comparability.} The two runtimes do not compute the same vanilla video: the shared runtime does not
forward frame groups that have finished denoising, the upstream runtime processes the whole valid interval. At the same
seed their vanilla outputs are different samples (SSIM $0.48$ between them on the $102$ fidelity prompts) of equal
quality (VBench aggregate difference $+0.004$ $[-0.023,+0.035]$ on those prompts), and the upstream vanilla takes
$589$\,s per video against $367$\,s. With every step recomputed, SCOPE's pipeline reproduces the upstream vanilla
exactly in the first window (checked on one prompt). We therefore (i) measure SCOPE's same-seed fidelity only against its own-stack vanilla,
generated for the $102$ fidelity prompts, (ii) normalize its speedup only to that vanilla, and (iii) compare across stacks
only what does not depend on the reference trajectory: latency in seconds on the same hardware and the VBench aggregate
on the same prompts. Fidelity columns of different stacks therefore measure closeness to different references and should not be
interpreted as a common ranking.

\emph{Results} (Table~\ref{tab:scope}). SCOPE is faster in seconds ($123$ vs.\ $161$\,s) and reproduces the
speedup its paper reports ($4.80\times$ here, $4.73\times$ there). Its VBench aggregate is $0.7748$,
$0.015$ below FSC-Quality ($[-0.024,-0.006]$) and $0.015$ below MotionCache ($[-0.024,-0.006]$) in the paired
bootstrap over the evaluated prompts; against its own vanilla on the fidelity prompts the difference is $-0.022$ $[-0.054,+0.007]$.
Its outputs deviate more from its own vanilla (SSIM $0.510$, PSNR $12.1$~dB, LPIPS
$0.570$) than FSC's and MotionCache's outputs deviate from theirs. The SCOPE paper reports higher fidelity
(SSIM $0.73$) under its own protocol ($257$ frames, a different prompt set); we did not investigate the difference beyond
the authors' own ablation: without selective computation (one flag of the released runner, reported in their Sec.~4.5) SCOPE
takes $155$\,s ($3.81\times$; their paper: $3.72\times$), is closer to its vanilla on $95$ of
$102$ prompts (SSIM $0.684$) and changes VBench by $+0.012$ $[-0.005,+0.028]$ on the $235$-prompt subset, where its
difference to FSC-Quality is $+0.001$ $[-0.015,+0.015]$. Because the protocols
differ substantially, we do not interpret this numerical gap as a reproduction failure.

\begin{table}[h]
\centering
\caption{\textbf{SCOPE is faster in seconds than FSC-Quality but lower on VBench; fidelity is not comparable
across runtime stacks.}
SCOPE on its official stack next to the shared-stack rows of Table~\ref{tab:main_results} (SkyReels-V2 1.3B, same
prompts, seed, generation settings and GPU model). \textbf{Speedup is normalized to each stack's own vanilla and fidelity is
measured against each stack's own same-seed vanilla ($102$ prompts)}; only latency and VBench are comparable across
stacks. $^\dagger$official ablation of the SCOPE paper, run on the $235$-prompt stratified subset and the fidelity prompts
($316$ videos); its VBench is the paired value on the subset's $126$ evaluated prompts, where the released configuration scores
$0.7769$ and FSC-Quality $0.7883$. Best in \textbf{bold}, second best \underline{underlined}, only in the
columns comparable across stacks (latency, VBench).}
\label{tab:scope}
\small
\setlength{\tabcolsep}{3pt}
\resizebox{\textwidth}{!}{%
\begin{tabular}{@{}llccccccc@{}}
\toprule
Runtime stack & Method & Own vanilla (s) & Latency (s)$\downarrow$ & Speedup (own vanilla)$\uparrow$ & VBench$\uparrow$ & PSNR$\uparrow$ & SSIM$\uparrow$ & LPIPS$\downarrow$ \\
\midrule
\multirow{2}{*}{shared (Table~\ref{tab:main_results})} & MotionCache & 367 & 161 & 2.275$\times$ & \underline{0.7898} & 17.34 & 0.7899 & 0.3366 \\
 & FSC-Quality (ours) & 367 & 161 & 2.275$\times$ & \textbf{0.7900} & 18.38 & 0.8209 & 0.3122 \\
\midrule
\multirow{2}{*}{official SCOPE (upstream)} & SCOPE, released configuration & 589 & \textbf{123} & 4.803$\times$ & 0.7748 & 12.10 & 0.5099 & 0.5696 \\
 & SCOPE, no selective computation$^\dagger$ & 589 & \underline{155} & 3.813$\times$ & 0.7889 & 14.37 & 0.6840 & 0.4380 \\
\bottomrule
\end{tabular}}
\end{table}

\subsection{High-Dynamics Stress Analysis}
\label{sec:appendix_stress}
Because VBench's per-video dynamic-degree score is coarse (the fraction of clips labelled dynamic), we score the
\emph{vanilla} videos with VBench's own RAFT pipeline and use the mean
top-$5\%$ flow magnitude (relative to VBench's motion threshold) as a
continuous motion measure, then split prompts into terciles.
On SkyReels-V2 (Table~\ref{tab:stress_sky}; $99$ seed-$7$ prompts with
retained videos: the $98$-prompt pool of App.~\ref{sec:appendix_seeds} plus one additional subject-consistency prompt;
$13$ of the $99$ are also in the $78$-prompt seed subset) FSC-Quality is ahead of MotionCache in every tercile, and the
margin \emph{grows} with motion (SSIM $+0.018$, $+0.035$, $+0.049$;
LPIPS $-0.004$, $-0.025$, $-0.051$).
FSC's motion attenuation
appears where vanilla motion is marginal (the mean dynamic-degree score drops by $0.06$
on low-motion prompts) and vanishes in the high-motion tercile.
On MAGI-1 ($110$ prompts; $88$ static and $22$ dynamic by that
label) FSC-Quality leads MotionCache on static prompts (SSIM $+0.040$, LPIPS
$-0.048$) and on dynamic ones (SSIM $+0.048$, LPIPS $-0.048$).

\begin{table}[h]
\centering
\caption{\textbf{FSC-Quality's fidelity margin over MotionCache grows with motion.}
SkyReels-V2 (FSC-Quality vs.\ MotionCache, MC), seed $7$, the $99$ retained prompts of App.~\ref{sec:appendix_stress} split using
vanilla-motion tercile boundaries defined on the original $100$-prompt pool (motion: mean top-$5\%$ RAFT flow magnitude relative to VBench's motion threshold).
Fidelity against the same-seed vanilla (offline metrics of Table~\ref{tab:seeds}, not on the scale of Table~\ref{tab:main_results}); $\Delta$DD = mean change of the
VBench dynamic-degree score (fraction of clips labelled dynamic) vs.\ vanilla. The better of the two methods in each fidelity metric is in \textbf{bold}.}
\label{tab:stress_sky}

\small
\setlength{\tabcolsep}{3pt}

\resizebox{\ifdim\width>\textwidth\textwidth\else\width\fi}{!}{%
\begin{tabular}{@{}lcc ccc ccc cc@{}}
\toprule
 & & & \multicolumn{3}{c}{MotionCache} & \multicolumn{3}{c}{FSC} & \multicolumn{2}{c}{$\Delta$DD} \\
Tercile & $n$ & motion & PSNR$\uparrow$ & SSIM$\uparrow$ & LPIPS$\downarrow$ & PSNR$\uparrow$ & SSIM$\uparrow$ & LPIPS$\downarrow$ & MC & FSC \\
\midrule
low & 34 & 0.60 & 18.05 & 0.6388 & 0.2864 & \textbf{19.13} & \textbf{0.6563} & \textbf{0.2820} & $+0.01$ & $-0.06$ \\
mid & 32 & 2.26 & 15.51 & 0.6102 & 0.3722 & \textbf{16.61} & \textbf{0.6448} & \textbf{0.3475} & $0.00$ & $-0.02$ \\
high & 33 & 6.70 & 14.41 & 0.5701 & 0.4402 & \textbf{15.80} & \textbf{0.6192} & \textbf{0.3893} & $-0.01$ & $0.00$ \\
\bottomrule
\end{tabular}%
}
\end{table}

\subsection{Motion Attenuation}
\label{sec:appendix_failure}
On MAGI-1, the VBench dimension on which FSC falls furthest behind
MotionCache is \emph{dynamic degree}.
There, FSC-Quality scores $0.295$ and FSC-Speed $0.236$ versus $0.365$ for
MotionCache and $0.434$ for FlowCache: with only $16$ denoising steps per window position
on this distilled backbone, reuse between scheduled full recomputations can damp
large motions, and this dimension alone accounts for FlowCache's higher VBench
aggregate on MAGI-1.
On SkyReels-V2, with $50$ denoising steps per temporal group and dense early anchors, the
effect vanishes in aggregate (App.~\ref{sec:appendix_stress}: FSC-Quality's dynamic-degree score is $0.06$ below vanilla on the
low-motion tercile and unchanged on the high-motion one): FSC-Quality and MotionCache both score $0.719$
and FSC-Speed $0.722$ (vanilla $0.740$).
On MAGI-1, highly dynamic prompts are therefore the regime in which more
conservative reuse is most important.

\subsection{No-Retuning Transfer to SkyReels-V2 14B}
\label{sec:appendix_14b}
To test cross-scale transfer, we applied the 1.3B FSC-Quality configuration (App.~\ref{sec:appendix_impl}) unchanged
to SkyReels-V2-DF-14B-540P, without additional hyperparameter search, keeping the same autoregressive settings, seed,
and baselines.
The 14B model's per-window key/value cache alone (${\sim}84$~GB
at 540P) does not fit on a single RTX PRO 6000 ($102.6$~GB) alongside the weights, so all four
methods are run at $352\times640$ with the runtime's CPU-offload setting, which offloads the text encoder and moves the
DiT between CPU and GPU across autoregressive windows; everything
else (177 frames, 97-frame window, \texttt{ar\_step} 5, 50 steps) is
identical to Table~\ref{tab:main_results}.
MotionCache and FlowCache use their released settings with the 14B
TeaCache polynomial.
Table~\ref{tab:sky14b} reports 47 prompts (five per benchmark prompt set, excluding the development prompts).
The VBench aggregate of FSC-Quality ($0.841$) is within the paired bootstrap CI of every other method (FSC-Quality$-$MotionCache
$+0.008$ $[-0.031,+0.036]$, FSC-Quality$-$FlowCache $+0.011$ $[-0.017,+0.042]$, FSC-Quality$-$vanilla $+0.009$ $[-0.014,+0.031]$).

\begin{table}[h]
\centering
\caption{\textbf{Unchanged from 1.3B, FSC-Quality is the fastest method on SkyReels-V2 14B, with the highest PSNR and
SSIM.} SkyReels-V2 14B, no retuning ($47$ prompts, $352\times640$,
seed $1024$; fidelity against the same-seed 14B vanilla; VBench on the $25$
prompts carrying the evaluated dimensions; memory in $10^9$ bytes on a $102.6$~GB card). Best accelerated result in \textbf{bold}, second
best \underline{underlined}.}
\label{tab:sky14b}

\small
\setlength{\tabcolsep}{3pt}

\resizebox{\ifdim\width>\textwidth\textwidth\else\width\fi}{!}{%
\begin{tabular}{@{}lcccccccc@{}}
\toprule
Method & Latency (s)$\downarrow$ & Speedup$\uparrow$ & Rel.\ FLOPs$\downarrow$ & Mem (GB)$\downarrow$ & VBench$\uparrow$ & PSNR$\uparrow$ & SSIM$\uparrow$ & LPIPS$\downarrow$ \\
\midrule
Vanilla     & 696 & 1.000$\times$ & 1.000 & 95.1 & 0.8313 & -- & -- & -- \\
FlowCache   & 437 & 1.595$\times$ & 0.581 & \textbf{96.7} & 0.8292 & \underline{16.07} & \underline{0.7692} & \textbf{0.3189} \\
MotionCache & \underline{389} & \underline{1.792$\times$} & \underline{0.504} & \underline{96.8} & \underline{0.8327} & 15.88 & 0.7600 & \underline{0.3289} \\
FSC-Quality (unchanged from 1.3B) & \textbf{328} & \textbf{2.121$\times$} & \textbf{0.392} & \underline{96.8} & \textbf{0.8406} & \textbf{16.18} & \textbf{0.7745} & 0.3302 \\
\bottomrule
\end{tabular}%
}
\end{table}

\subsection{Further Separation Variants and \texorpdfstring{$2\times2$}{2x2} Factorial}
\label{sec:appendix_separation}
Table~\ref{tab:separation} lists the variants of the separation study that are not in Table~\ref{tab:ablation},
on the same development set at matched wall-clock.
The $2\times2$ factorial crosses the anchor schedule (uniform vs.\
dense-to-sparse) with the repair controller (off vs.\ on); anchor counts
are chosen so that every cell runs within about $5\%$ of FSC-Quality's latency (the other three cells are rows of
Table~\ref{tab:ablation}).
Both levels contribute: dense-to-sparse placement adds
$0.017$ SSIM over uniform anchors without repair, the repair controller adds
$0.068$ over uniform anchors alone, and the combination adds $0.088$.
Phase shaping mainly affects latency ($<0.002$ SSIM, $3\%$ latency).

\begin{table}[h]
\centering
\caption{
\textbf{Further variants of the separation study: uniform anchors without repair are outperformed on all eight prompts, while
removing phase shaping changes SSIM by at most $0.002$.}
SkyReels-V2, our development set (eight prompts), seed $1024$, median wall-clock per video on one idle GPU; rows modify
FSC-Quality (D2S: dense-to-sparse); the purely reactive rows and the other factorial cells are in
Table~\ref{tab:ablation}.
``Collapsed repair thresholds'' collapses the $C1$--$C3$ risk thresholds to a single level; safeguard-triggered repairs
retain their native repair budget (this is not a fixed-budget controller). The last two columns count the prompts on which
FSC-Quality outperforms the variant. Best in \textbf{bold}, second best \underline{underlined}.
}
\label{tab:separation}

\small
\setlength{\tabcolsep}{3pt}

\resizebox{\ifdim\width>\textwidth\textwidth\else\width\fi}{!}{%
\begin{tabular}{@{}lcccccc@{}}
\toprule
Variant & Anchors & s/video~$\downarrow$ & SSIM~$\uparrow$ & LPIPS~$\downarrow$ & \multicolumn{2}{c}{FSC-Quality wins (of 8)} \\
\cmidrule(lr){6-7}
 & & & & & SSIM & LPIPS \\
\midrule
FSC-Quality: repair controller, D2S anchors & 10 D2S & \underline{161} & \underline{0.6641} & \underline{0.2925} & -- & -- \\
\midrule
uniform anchors only (no repair)          & 13 uniform & \textbf{153} & 0.5766 & 0.3496 & 8 & 8 \\
collapsed repair thresholds               & 10 D2S     & 165 & 0.6596 & 0.3024 & 5 & 4 \\
no phase shaping                          & 10 D2S     & 166 & \textbf{0.6658} & \textbf{0.2916} & 2 & 2 \\
\bottomrule
\end{tabular}%
}
\end{table}

\subsection{Anchor Placement: Influence Curve and Schedule Search}
\label{sec:appendix_anchor_search}
Two experiments assess the empirical basis for the dense-to-sparse schedule of
Eq.~\ref{eq:anchors}.

\begin{figure}[H]
\centering
\includegraphics[width=\linewidth]{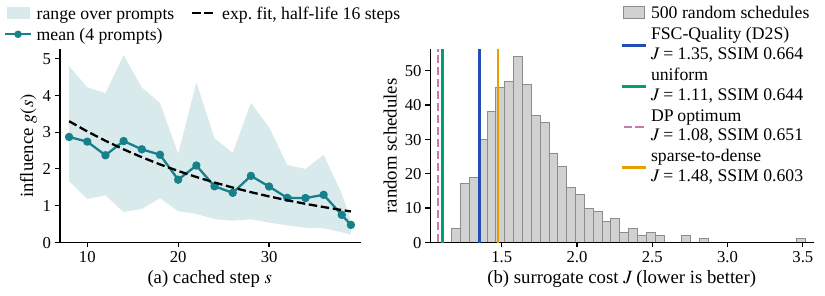}
\caption{\textbf{The influence of a cached step decays along the trajectory, but a local-error surrogate does
not select the end-to-end best schedule.}
\textbf{(a)} Influence $g(s)$ of a single cached step on the final video
(final LPIPS damage per unit local reuse error; mean and range over four
profiling prompts, $72$ generations): influence decays about $4$--$6\times$ along
the trajectory (endpoint ratio $6.1\times$).
\textbf{(b)} Local-error surrogate cost of $500$ random ten-anchor
schedules (histogram) and of the hand-crafted, uniform, reversed, and
unweighted DP-optimal schedules (lines), each annotated with its end-to-end
SSIM on our development set (Table~\ref{tab:anchor_search}); the random schedules are scored under
the surrogate only.}
\label{fig:influence_curve}
\end{figure}

\paragraph{Influence of a single cached step.}
For each step $s\in\{6,8,\ldots,38,39\}$ and four profiling prompts we
replay step $s$ from the cache with every other step exact ($72$
generations) and divide the resulting damage in the final video (LPIPS to
the same-seed vanilla) by the teacher-forced local reuse error of that step,
giving the influence $g(s)$ of a unit local error
(Fig.~\ref{fig:influence_curve}(a)).
$g(s)$ shows a strong decreasing trend along the trajectory (Spearman
$-0.95$ with $s$; exponential-fit half-life $16$ steps; $6.1\times$ from step $8$
to $39$,
$3.8\times$ and $2.9\times$ when damage is measured by $1{-}$SSIM or relative
$L_2$).
Local errors are therefore not interchangeable: an early one is amplified
several times more than a late one, which is the mechanism behind
Observation~1 and the reason anchor gaps should grow along the trajectory;
it says nothing about \emph{which} growing schedule to use.

\paragraph{Schedule search.}
We next evaluate whether a local surrogate can recover the schedule.
From instrumented full-compute traces of eight profiling prompts we
measure, for every candidate anchor $i$ and later step $s$, the mismatch
between the cached whole-step residual of $i$ and the exact residual of
$s$, and minimize the summed mismatch over ten anchors by dynamic
programming, with and without weighting each term by $g(s)$; we also draw
$500$ random ten-anchor schedules.
The hand-crafted schedule sits in the best $10\%$ of the random schedules
but is not the surrogate optimum: the unweighted optimum is nearly uniform
(gaps $3,3,3,4,4,4,4,4,4$), and the $g$-weighted optima are dense-to-sparse
in shape (gaps $2,3,3,3,4,4,4,5,5$ and $2,2,3,3,4,4,5,5,5$).
End-to-end, however, every one of them is outperformed by the hand-crafted schedule
with the rest of FSC-Quality held fixed on our development set (Table~\ref{tab:anchor_search};
Fig.~\ref{fig:influence_curve}(b)), and the
surrogate mis-orders the schedules it can compare (it ranks uniform ahead of
dense-to-sparse).
The surrogate-optimal schedules run $2.6\%$ faster; this latency difference can account for at most about $0.008$
SSIM of their $0.008$--$0.020$ deficit.
The influence curve therefore motivates growing gaps; the anchor locations themselves are selected empirically.

\begin{table}[h]
\centering
\caption{\textbf{Every alternative anchor schedule has lower mean SSIM than FSC-Quality's dense-to-sparse schedule.}
Anchor schedules with the rest of FSC-Quality held fixed (SkyReels-V2,
our development set of eight prompts, seed $1024$, ten anchors each; median s/video).
$J$ = local-error surrogate cost (lower is better under the surrogate);
DP = surrogate optimum, unweighted or weighted by the measured influence
$g(s)$. The last two columns count the prompts on which the hand-crafted
schedule wins. The FSC-Quality, uniform, and reversed rows are the runs of
Table~\ref{tab:ablation}; the DP schedules were generated separately with
identical code, seed, and settings, and their win counts are paired against
that FSC-Quality run. Best in \textbf{bold}, second best \underline{underlined}.}
\label{tab:anchor_search}

\small
\setlength{\tabcolsep}{3pt}

\resizebox{\ifdim\width>\textwidth\textwidth\else\width\fi}{!}{%
\begin{tabular}{@{}llcccccc@{}}
\toprule
Schedule & Gaps & $J\downarrow$ & s/video$\downarrow$ & SSIM~$\uparrow$ & LPIPS~$\downarrow$ & \multicolumn{2}{c}{FSC-Quality wins (of 8)} \\
\cmidrule(lr){7-8}
 & & & & & & SSIM & LPIPS \\
\midrule
FSC-Quality: dense-to-sparse (Eq.~\ref{eq:anchors}) & 1,1,2,3,4,4,6,5,7 & 1.353 & \underline{161} & \textbf{0.6641} & \textbf{0.2925} & -- & -- \\
DP, $g$-weighted ($1{-}$SSIM) & 2,3,3,3,4,4,4,5,5 & 1.126 & \textbf{157} & \underline{0.6558} & \underline{0.3019} & 5 & 5 \\
DP, unweighted                & 3,3,3,4,4,4,4,4,4 & \textbf{1.077} & \textbf{157} & 0.6510 & 0.3104 & 5 & 5 \\
DP, $g$-weighted (LPIPS)      & 2,2,3,3,4,4,5,5,5 & 1.146 & \textbf{157} & 0.6442 & 0.3197 & 6 & 6 \\
uniform                       & 4,3,4,4,3,4,4,3,4 & \underline{1.108} & \textbf{157} & 0.6441 & 0.3189 & 5 & 5 \\
sparse-to-dense (reversed)    & 7,5,6,4,4,3,2,1,1 & 1.476 & \underline{161} & 0.6033 & 0.3659 & 8 & 8 \\
\bottomrule
\end{tabular}%
}
\end{table}

\subsection{Local Hyperparameter Sensitivity}
\label{sec:appendix_sensitivity}
\begin{figure}[t]
\centering
\includegraphics{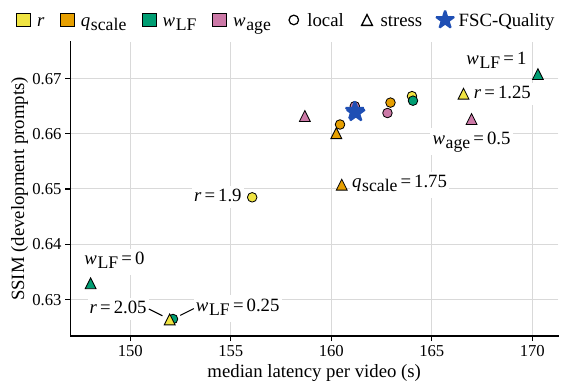}
\caption{\textbf{Perturbations toward more repair are marginally better at slightly higher latency, and the larger
deviations toward less repair trade SSIM for lower latency.} SSIM against median latency per video around FSC-Quality
(SkyReels-V2, our development set; stars: the two FSC-Quality runs; circles: local perturbations; triangles: stress
perturbations; color: perturbed parameter; values in Table~\ref{tab:sensitivity}).}
\label{fig:sensitivity}
\end{figure}

Around FSC-Quality we vary one parameter at a time, with every other parameter held fixed: the
threshold scale $r$ ($\pm15\%$), $q_{\mathrm{scale}}$ ($\pm20\%$), and the relative weights $w_{\mathrm{LF}}/w_{\mathrm{HF}}$
and $w_{\mathrm{age}}/w_{\mathrm{HF}}$ ($\pm50\%$, with $w_{\mathrm{HF}}{=}1$ as the normalization), plus a wider stress
endpoint on each axis. All $144$ runs use our development set and one idle
GPU; FSC-Quality is run twice (both runs take the same routing decisions; mean $|\Delta\mathrm{SSIM}|{=}0.0008$ between
the two runs). Table~\ref{tab:sensitivity} lists every point. Six of the eight local points change SSIM by at most
$0.0028$, LPIPS by at most $0.0056$ and latency by at most $1.8\%$; over all eight local
points the maxima are $0.0375$, $0.0503$ and $5.6\%$.
The two larger local deviations ($r{=}1.90$ and $w_{\mathrm{LF}}{=}0.25$) trade $0.015$--$0.038$ SSIM for $3$--$6\%$ lower latency.
Perturbations toward more repair are marginally better at slightly higher latency ($r{=}1.40$:
$+0.0028$ SSIM at $+1.7\%$ latency; $w_{\mathrm{LF}}{=}0.75$:
$+0.0020$ at $+1.8\%$; Fig.~\ref{fig:sensitivity}).

\begin{table}[t]
\centering
\caption{\textbf{Six of the eight local perturbations of FSC-Quality change SSIM by at most $0.0028$ and latency by at
most $1.8\%$.} Local sensitivity around FSC-Quality (SkyReels-V2, our development set of eight prompts): $\Delta$
relative to the mean of the two FSC-Quality runs, routing shares over non-anchor decisions, and W/L = prompts on which
the point outperforms/underperforms FSC-Quality; local: $r\pm15\%$, $q_{\mathrm{scale}}\pm20\%$, relative weights $\pm50\%$; stress: a wider endpoint on each axis; routing levels: $C0$ whole-step replay, $C1$--$C3$ token repair of increasing fraction, Full full repair (Sec.~\ref{sec:framework_repair}). s/video: median wall-clock. Best perturbed point in \textbf{bold}, second best \underline{underlined}.}
\label{tab:sensitivity}
\resizebox{\linewidth}{!}{%
\begin{tabular}{llrrrrrrrccc}
\toprule
Point & Role & s/video$\downarrow$ & $\Delta$lat.\,\% & PSNR$\uparrow$ & SSIM$\uparrow$ & $\Delta$SSIM & LPIPS$\downarrow$ & $\Delta$LPIPS & C0/C1/C2/C3/Full\,\% & SSIM W/L & LPIPS W/L \\
\midrule
FSC-Quality (run 1) & -- & 161 & $+0.0$ & 18.92 & 0.6641 & $+0.0002$ & 0.2926 & $-0.0002$ & 70/12/3/9/6 & 7/1 & 5/3 \\
FSC-Quality (run 2) & -- & 161 & $+0.0$ & 18.90 & 0.6638 & $-0.0002$ & 0.2931 & $+0.0002$ & 70/12/3/9/6 & 1/7 & 3/5 \\
$r$$=1.4$ & local & 164 & $+1.7$ & 18.91 & 0.6668 & $+0.0028$ & 0.2905 & $-0.0024$ & 68/10/4/12/6 & 7/1 & 5/3 \\
$r$$=1.9$ & local & 156 & $-3.2$ & 18.76 & 0.6485 & $-0.0155$ & 0.3172 & $+0.0243$ & 71/12/4/11/3 & 1/7 & 1/7 \\
$q_{\mathrm{scale}}$$=1$ & local & 160 & $-0.5$ & 18.83 & 0.6617 & $-0.0023$ & 0.2985 & $+0.0056$ & 70/12/3/9/6 & 2/6 & 2/6 \\
$q_{\mathrm{scale}}$$=1.5$ & local & 163 & $+1.1$ & 18.89 & 0.6656 & $+0.0017$ & 0.2887 & $-0.0042$ & 70/12/3/9/6 & 7/1 & 6/2 \\
$w_{\mathrm{LF}}$$=0.25$ & local & \underline{152} & $-5.6$ & 18.43 & 0.6265 & $-0.0375$ & 0.3432 & $+0.0503$ & 71/14/3/12/0 & 0/8 & 0/8 \\
$w_{\mathrm{LF}}$$=0.75$ & local & 164 & $+1.8$ & 18.89 & 0.6660 & $+0.0020$ & 0.2906 & $-0.0023$ & 67/10/5/12/6 & 7/1 & 6/2 \\
$w_{\mathrm{age}}$$=0.125$ & local & 161 & $+0.0$ & \underline{18.92} & 0.6650 & $+0.0010$ & 0.2920 & $-0.0009$ & 71/12/3/9/6 & 6/2 & 5/3 \\
$w_{\mathrm{age}}$$=0.375$ & local & 163 & $+1.0$ & 18.89 & 0.6637 & $-0.0002$ & 0.2938 & $+0.0009$ & 68/12/4/10/6 & 3/5 & 3/5 \\
$r$$=1.25$ & stress & 167 & $+3.3$ & 18.89 & \underline{0.6672} & $+0.0033$ & \underline{0.2844} & $-0.0085$ & 65/11/6/10/7 & 6/2 & 6/2 \\
$r$$=2.05$ & stress & \underline{152} & $-5.7$ & 18.46 & 0.6264 & $-0.0376$ & 0.3426 & $+0.0497$ & 71/13/4/12/0 & 0/8 & 0/8 \\
$q_{\mathrm{scale}}$$=0.75$ & stress & 160 & $-0.6$ & 18.81 & 0.6601 & $-0.0039$ & 0.3002 & $+0.0073$ & 70/12/3/9/6 & 1/7 & 0/8 \\
$q_{\mathrm{scale}}$$=1.75$ & stress & 161 & $-0.4$ & 18.65 & 0.6508 & $-0.0132$ & 0.3268 & $+0.0339$ & 70/12/3/12/3 & 0/8 & 0/8 \\
$w_{\mathrm{LF}}$$=0$ & stress & \textbf{148} & $-8.2$ & 18.53 & 0.6329 & $-0.0310$ & 0.3368 & $+0.0439$ & 76/12/5/6/0 & 0/8 & 0/8 \\
$w_{\mathrm{LF}}$$=1$ & stress & 170 & $+5.6$ & \textbf{18.94} & \textbf{0.6708} & $+0.0068$ & \textbf{0.2757} & $-0.0172$ & 65/11/5/10/9 & 6/2 & 6/2 \\
$w_{\mathrm{age}}$$=0$ & stress & 159 & $-1.6$ & 18.87 & 0.6632 & $-0.0008$ & 0.2942 & $+0.0013$ & 74/10/4/7/6 & 3/5 & 2/6 \\
$w_{\mathrm{age}}$$=0.5$ & stress & 167 & $+3.6$ & 18.84 & 0.6626 & $-0.0013$ & 0.2952 & $+0.0023$ & 62/14/6/12/6 & 4/4 & 2/6 \\
\bottomrule
\end{tabular}}
\end{table}

\end{document}